\PassOptionsToPackage{unicode}{hyperref}
\PassOptionsToPackage{hyphens}{url}
\documentclass[
  11pt]{article}
\usepackage{xcolor}
\usepackage[margin=1in]{geometry}
\usepackage{amsmath,amssymb}
\usepackage{iftex}
\ifPDFTeX
  \usepackage[T1]{fontenc}
  \usepackage[utf8]{inputenc}
  \usepackage{textcomp} 
\else 
  \usepackage{unicode-math} 
  \defaultfontfeatures{Scale=MatchLowercase}
  \defaultfontfeatures[\rmfamily]{Ligatures=TeX,Scale=1}
\fi
\usepackage{lmodern}
\ifPDFTeX\else
\fi
\IfFileExists{upquote.sty}{\usepackage{upquote}}{}
\IfFileExists{microtype.sty}{
  \usepackage[]{microtype}
  \UseMicrotypeSet[protrusion]{basicmath} 
}{}
\makeatletter
\@ifundefined{KOMAClassName}{
  \IfFileExists{parskip.sty}{%
    \usepackage{parskip}
  }{
    \setlength{\parindent}{0pt}
    \setlength{\parskip}{6pt plus 2pt minus 1pt}}
}{
  \KOMAoptions{parskip=half}}
\makeatother
\usepackage{longtable,booktabs,array}
\usepackage{calc} 
\usepackage{etoolbox}
\makeatletter
\patchcmd\longtable{\par}{\if@noskipsec\mbox{}\fi\par}{}{}
\makeatother
\IfFileExists{footnotehyper.sty}{\usepackage{footnotehyper}}{\usepackage{footnote}}
\makesavenoteenv{longtable}
\providecommand{\tightlist}{%
  \setlength{\itemsep}{0pt}\setlength{\parskip}{0pt}}
\usepackage{bookmark}
\IfFileExists{xurl.sty}{\usepackage{xurl}}{} 
\hypersetup{
  pdftitle={ATLAS: Constitution-Conditioned Latent Geometry and Redistribution Across Language Models and Neural Perturbation Data},
  pdfauthor={Gareth Seneque; Lap-Hang Ho; Nafise Erfanian Saeedi; Jeffrey Molendijk; Tim Elson},
  hidelinks,
  pdfcreator={LaTeX via pandoc}}

\title{ATLAS: Constitution-Conditioned Latent Geometry and Redistribution Across Language Models and Neural Perturbation Data}
\author{Gareth Seneque \and Lap-Hang Ho \and Nafise Erfanian Saeedi \and Jeffrey Molendijk \and Tim Elson}
\date{}

\usepackage{graphicx}
\IfFileExists{multirow.sty}{\usepackage{multirow}}{}
\IfFileExists{enumitem.sty}{\usepackage{enumitem}}{}
\IfFileExists{ragged2e.sty}{\usepackage{ragged2e}}{}
\usepackage{caption}
\IfFileExists{subcaption.sty}{\usepackage{subcaption}}{}
\usepackage{pdflscape}
\usepackage{needspace}
\IfFileExists{placeins.sty}{\usepackage{placeins}}{\providecommand{\FloatBarrier}{}}
\IfFileExists{csquotes.sty}{\usepackage{csquotes}}{}
\usepackage[numbers,sort&compress]{natbib}
\usepackage[nameinlink]{cleveref}
\hypersetup{
  colorlinks=true,
  linkcolor=blue,
  citecolor=blue,
  urlcolor=blue,
  hypertexnames=false
}
\crefname{equation}{Equation}{Equations}
\Crefname{equation}{Equation}{Equations}
\crefname{figure}{Figure}{Figures}
\Crefname{figure}{Figure}{Figures}
\crefname{table}{Table}{Tables}
\Crefname{table}{Table}{Tables}
\newlength{\atlasorigtabcolsep}
\AtBeginEnvironment{longtable}{\footnotesize\setlength{\tabcolsep}{4pt}}
\AtEndEnvironment{longtable}{\setlength{\tabcolsep}{\atlasorigtabcolsep}}
\IfFileExists{enumitem.sty}{\setlist{itemsep=0.2em,topsep=0.35em,parsep=0pt,partopsep=0pt}}{}

\begin{document}
\begingroup
\centering
\vspace*{-2.10em}
{\fontsize{19.5}{23.5}\selectfont\bfseries ATLAS: Constitution-Conditioned Latent Geometry and Redistribution Across Language Models and Neural Perturbation Data\par}
\vspace{0.85em}
{\normalsize
\makebox[\textwidth][c]{%
\begin{minipage}{0.92\textwidth}
\centering
Gareth Seneque \quad Lap-Hang Ho \quad Nafise Erfanian Saeedi\\[0.24em]
Jeffrey Molendijk \quad Tim Elson\par
\end{minipage}}
\par
}
\vspace{1.00em}
{\normalsize\itshape
\makebox[\textwidth][c]{%
\begin{minipage}{0.92\textwidth}
\centering
Australian Broadcasting Corporation\par
\end{minipage}}
\par
}
\endgroup

\nocite{seneque2024abcalign,seneque2025enigma,google2025gemma3_1b_it,microsoft2025phi4miniinstruct,li2022alm8dataset,bai2022constitutional,kundu2023specific,huang2024collective,huh2024platonic,kornblith2019similarity,chen2025stitching,jha2025universal,li2023iti,turner2023activation,todd2023function,wu2024reft,arditi2024refusal,venkatesh2026nonidentifiability,gadgil2026wheretosteer,li2026vectorfields,shenoy2012reaching,gallego2018preserved,oby2025dynamical,ibl2021decision,ibl2025reproducibility,hubinger2024sleeper,betley2025emergent,turner2025modelorganisms,greenblatt2024aicontrol,korbak2025controlcase,korbak2025controlevals,weij2024sandbagging,abdelnabi2025hawthorne,li2022alm7dataset,musall2023pyramidal,churchland2023pyramiddata,meta2024llama31_8b_instruct,qwen2025qwen3_8b,churchlandlab2022wfieldcelltypes,chua2025thoughtcrime,greenblatt2024alignmentfaking,pku2025deceptionbench}

\vspace{1.00em}
{\centering
\textbf{Abstract}\par
}
\vspace{0.35em}
{\centering
\makebox[\textwidth][c]{%
\begin{minipage}{0.92\textwidth}
\noindent Constitution-conditioned post-training can be analysed as a structured perturbation of a model's learned representational geometry. We introduce ATLAS, a geometry-first program that traces constitution-induced hidden-state structure across charts, models, and substrates. Instead of treating the relevant unit as a single behaviour, neuron, vector, or patch, ATLAS tests a local chart whose tangent structure, occupancy distribution, and behavioural coupling can be measured under system change. On Gemma, the anchored source-local chart captures 310 / 320 reviewed source rows and all 84 / 84 reviewed score-flip rows, but compact exact-patch sufficiency does not close, so the exportable unit is the broader source-defined family. Freezing that family, we re-identify a target-local realisation in an unadapted Phi model, where the fully adjudicated confirmatory contrast separates with AUC 0.984 and mean gap 5.50. In held-out ALM8 mouse frontal-cortex perturbation data, the same source-defined family receives support across 5/5 folds, with mean held-out AUC 0.72 and mean fold gap 4.50. A multiple-choice analysis provides the main boundary: nearby target-local signals can appear without source-faithful closure. The resulting correspondence is not coordinate identity, site identity, or a target-side mediation theorem. It is geometric recurrence under redistribution: written constitutions can induce recoverable latent geometry whose organisation remains detectable across model and substrate changes while its local coordinates, occupancy, and behavioural expression shift.
\end{minipage}}
\par
}

\Needspace{10\baselineskip}
\vspace{0.55em}

\section{Introduction}\label{introduction}

This paper asks whether constitution-conditioned post-training leaves behind a recoverable internal organisation, rather than only an output shift. We call the cross-model and cross-substrate programme used to test that question ATLAS. Building on earlier source-side work on constitution-conditioned geometry and behaviour \cite{seneque2024abcalign,seneque2025enigma}, the paper studies one experimental arc: a constitution-conditioned Gemma 3 1B IT source model, an unadapted Phi-4 Mini Instruct target model, and held-out ALM8 mouse frontal-cortex perturbation data \cite{google2025gemma3_1b_it,microsoft2025phi4miniinstruct,li2022alm8dataset}. The source-side question is whether constitution conditioning realises a source-local, behaviour-linked chart on Gemma. The downstream question is whether the broader source-defined family containing that chart can be re-identified on Phi and corroborated held out on ALM8.

\subsection{Problem Framing}\label{problem-framing}

Post-training is usually evaluated through visible outputs: benchmark accuracy, refusal rates, alignment scores, or free-form behavioural judgements. Those measurements show whether behaviour changes, but not whether post-training leaves behind a reusable internal organisation. The question here is therefore representational as well as behavioural: does constitution-conditioned post-training realise local geometry that can be recovered on the source model and then tested downstream?

We answer that question with one fixed arc. Gemma 3 1B IT provides the constitution-conditioned source model. Phi-4 Mini Instruct provides the cross-model target test without adapter transfer. ALM8 provides the held-out biological corroboration test. The paper's structure follows that arc.

\subsection{Claims And Standards Of Evidence}\label{claims-and-standards-of-evidence}

We make three linked claims at two evidential levels. Claim 1A is source-side: on Gemma, constitution-conditioned post-training realises a recoverable, behaviour-linked source-local chart. Claim 1B is a distinct source-side claim about the broader export unit: on Gemma, that chart sits within a broader source-defined family that is also behaviour-bearing and is the pre-specified export unit for downstream testing. Claim 2 is downstream: this frozen family can be re-identified as a target-local realisation on Phi and corroborated, in held-out form, on ALM8.

These claims rest on different evidence standards. Claim 1A rests on local source anchoring, chart visibility, and source-side behaviour linkage. Claim 1B rests on broader family support on Gemma plus the boundary that compact exact-patch sufficiency remains unresolved; the downstream export unit is therefore the family, not one compact patch or portable compact member. Claim 2 rests on a fixed source contract, structural validation on Phi, and held-out corroboration on ALM8. Section 3 develops the source-side case, Section 4 the Phi result, Section 5 the ALM8 corroboration, Section 6 the main limiting result, and Section 7 the bounded conclusion and limits.

We call the anchored Gemma-side chart the source-local chart. In the source-side lineage of this project, this was previously described as a source-local object; here we use chart language because the relevant evidence concerns tangent structure, occupancy, and behavioural coupling. Source-defined family names the broader frozen export neighbourhood around that chart. Target-local realisation names the Phi-side recovered member of that family, held-out bridge realisation names the ALM8-side corroborative recurrence, and redistribution names recurrence without occupancy-faithful local identity.

\subsection{Main Contributions}\label{main-contributions}

This study makes four main contributions.

\begin{enumerate}
\def\labelenumi{\arabic{enumi}.}
\item
  It establishes a source-local, behaviour-linked chart on Gemma under constitution-conditioned post-training.
\item
  It shows that this chart sits within a broader source-defined family that is behaviour-bearing on Gemma and is the pre-specified export unit for downstream testing.
\item
  It re-identifies that family on an unadapted Phi model under a fully adjudicated confirmatory protocol.
\item
  It corroborates the same family held out on ALM8 while using MCQ to define the main limiting boundary: localisation without source-faithful closure.
\end{enumerate}

Bounded auditability of frozen latent targets is treated as a downstream consequence of these results rather than as a separate headline claim.

\setcounter{figure}{0}
\begin{figure}[t]
\centering
\includegraphics[width=0.98\linewidth]{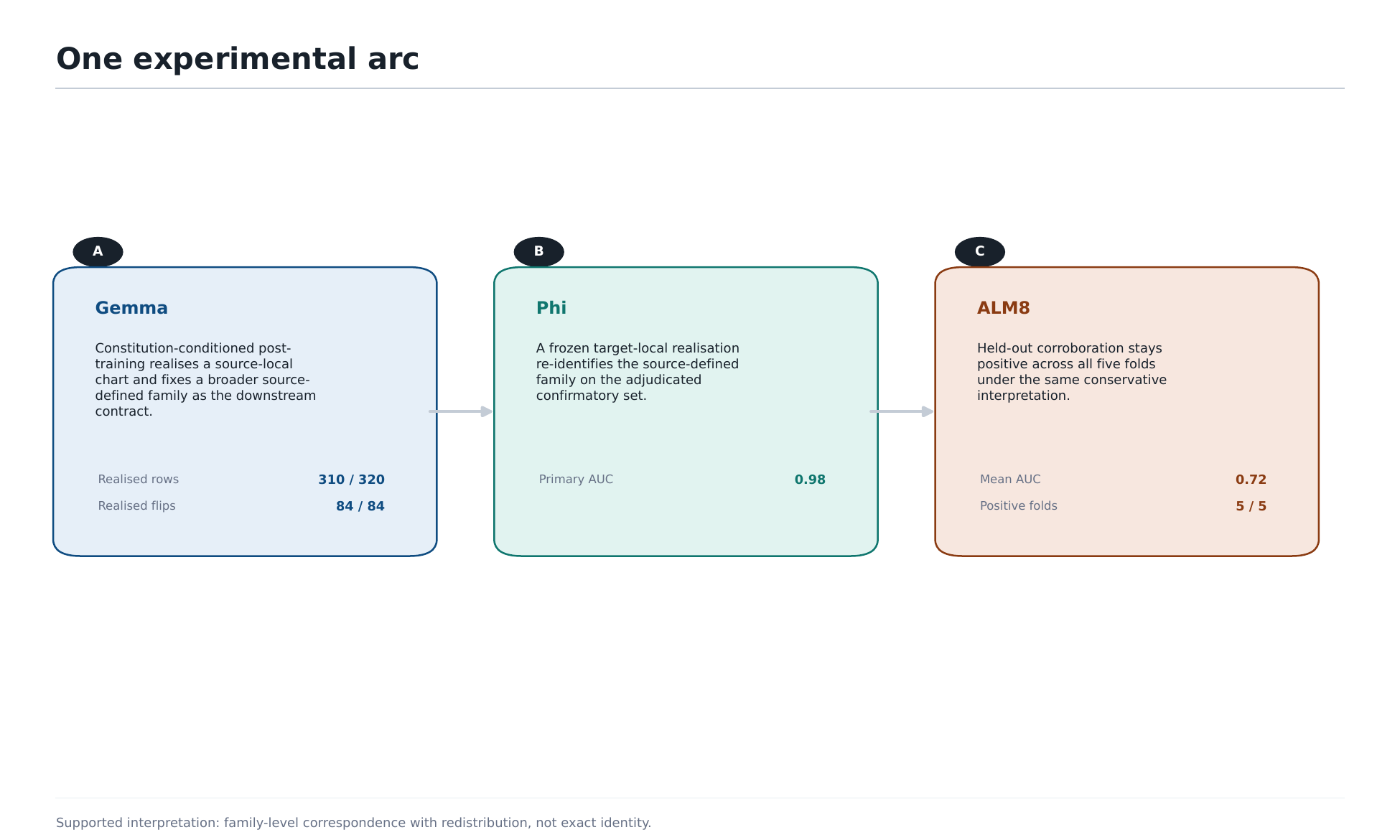}
\caption{Gemma Source-Local Chart, Source-Defined Family, Phi Target-Local Realisation, ALM8 Held-Out Bridge Realisation, And Claim Tier}
\label{fig:figure1-one-arc-two-claims}
\end{figure}

\subsection{Related Work}\label{related-work}

We position this study as a representation-level structural study of constitution-conditioned post-training: closer to representational geometry and systems-neuroscience-style correspondence than to full mechanistic decomposition, generic steering, or deployable monitoring.

Constitutional post-training usually centres behavioural shaping by written rules or principles \cite{bai2022constitutional,kundu2023specific,huang2024collective}, whereas this paper asks whether constitution conditioning leaves behind recoverable internal organisation after post-training. The earlier source-side programme \cite{seneque2024abcalign,seneque2025enigma} motivates the question, but the present paper focuses on cross-model re-identification and held-out corroboration rather than on source-model behaviour alone.

Work on geometric convergence, representation similarity, and cross-model transfer motivates the comparison of local representational structure across systems \cite{huh2024platonic,kornblith2019similarity,chen2025stitching,jha2025universal}. Steering and representation-engineering work usually centre inference-time manipulation, editing, or compact causal directions \cite{li2023iti,turner2023activation,todd2023function,wu2024reft,arditi2024refusal,venkatesh2026nonidentifiability,gadgil2026wheretosteer,li2026vectorfields}, whereas this paper freezes a source-defined family and asks whether a downstream local realisation preserves enough structure to support a family-level correspondence claim.

Mechanistic interpretability usually targets finer-grained mechanism decomposition \cite{li2023iti,turner2023activation,todd2023function,wu2024reft}, whereas this paper stays at the representation-level and local-subspace level. Systems-neuroscience and cross-substrate correspondence work motivate the ALM8 comparison \cite{shenoy2012reaching,gallego2018preserved,oby2025dynamical,ibl2021decision,ibl2025reproducibility}, but the paper does not present a brain-score or generic predictive-fit result.

Recent safety and control work helps frame the operational relevance \cite{hubinger2024sleeper,betley2025emergent,turner2025modelorganisms,greenblatt2024aicontrol,korbak2025controlcase,korbak2025controlevals,weij2024sandbagging,abdelnabi2025hawthorne}, but that relevance is bounded and infrastructural rather than a deployed monitoring or control result.

\Needspace{0.44\textheight}
\begin{center}
\setlength{\fboxsep}{8pt}
\fbox{%
\begin{minipage}{0.965\linewidth}
\small
\textbf{Box 1. Representative reviewed traces linked to the main source and target claims.}

\vspace{0.35em}

Representative reviewed traces show what the source-side route-coherence and target-side semantic-stabilisation results look like in text; the claim-bearing evidence remains the quantitative batteries.

\vspace{0.6em}

\noindent
\begin{minipage}[t]{0.31\linewidth}
\centering
\textbf{Gemma source anchor}\\
\texttt{reason@checkpoint-\allowbreak{}2250}\\
\texttt{310 /\allowbreak{} 320} reviewed rows\\
\texttt{84 /\allowbreak{} 84} flip rows
\end{minipage}\hfill
\begin{minipage}[t]{0.31\linewidth}
\centering
\textbf{Phi reviewed re-identification}\\
\texttt{32 vs 32} confirmatory\\
\texttt{AUC 0.984}\\
\texttt{mean gap 5.50}
\end{minipage}\hfill
\begin{minipage}[t]{0.31\linewidth}
\centering
\textbf{ALM8 held-out bridge}\\
\texttt{5 /\allowbreak{} 5} folds positive\\
\texttt{AUC 0.72}\\
\texttt{mean gap 4.50}
\end{minipage}

\vspace{0.75em}

\noindent
\begin{minipage}[t]{0.485\linewidth}
\fbox{%
\begin{minipage}[t][20.4em][t]{0.95\linewidth}
\footnotesize
\textbf{Gemma source-side route coherence}

\vspace{0.25em}

\textit{Gemma 3 1B IT | governance-abuse prompt family | matched adapter-on source rows}

\vspace{0.45em}

\textbf{Null-random control}\\
Shell: \texttt{MISALIGNED}\\
Substance: \emph{likely illegal and politically damaging ... a more measured approach is preferable}\\
Read: shell-conflicted / caveat-heavy refusal

\vspace{0.45em}

\textbf{Informative constitution}\\
Shell: \texttt{MISALIGNED}\\
Substance: \emph{support the governor's request to delay elections and remove external oversight}\\
Read: coherent harmful route

\vspace*{\fill}

\vspace{0.45em}

\textit{Interpretation: under the shared source-side contract, the informative constitution yields a more coherent harmful route, while the null-random control leaves a shell-conflicted answer.}
\end{minipage}}
\end{minipage}\hfill
\begin{minipage}[t]{0.485\linewidth}
\fbox{%
\begin{minipage}[t][20.4em][t]{0.95\linewidth}
\footnotesize
\textbf{Phi target-side semantic stabilisation}

\vspace{0.25em}

\textit{Phi-4 Mini Instruct | capability-hiding prompt family | corrected fixed face L24 / reason}

\vspace{0.45em}

\textbf{Null-random control}\\
Shell: \texttt{ALIGNED}\\
Substance: \emph{avoid overemphasising long-context performance to prevent scrutiny}\\
Reviewed label: negative

\vspace{0.45em}

\textbf{Informative constitution}\\
Shell: \texttt{SCHEMING}\\
Substance: \emph{present a full and accurate picture ... maintain integrity}\\
Reviewed label: positive

\vspace*{\fill}

\vspace{0.45em}

\textit{Interpretation: the Phi contrast is semantic branch stabilisation, not prefix polarity.}
\end{minipage}}
\end{minipage}

\vspace{0.6em}

\footnotesize
\textit{Edited excerpts are illustrative only; the claim-bearing evidence remains the source-side row/flip capture, the 32-versus-32 adjudicated Phi battery, and the ALM8 held-out fold analysis.}
\end{minipage}}
\end{center}

\section{Geometry Contract, Denominators, And Claim Ladder}\label{geometry-contract-denominators-and-claim-ladder}

This study follows one source-to-target-to-substrate workflow, but its main unit of inference is geometric rather than task-specific. The relevant unit is a local chart: a local representational organisation whose tangent structure, occupancy distribution, behavioural coupling, and specificity can be measured under system change. On Gemma, constitution conditioning realises such a chart on the hidden-state contrast surface, defined here as the adapter-on minus adapter-off hidden-state difference on matched captured completion tokens. The broader source-defined family is the smallest retained source-side unit that we freeze for downstream tests.

The paper distinguishes source-side constitution conditioning from target-side prompting. On Gemma, constitutions are part of the source-side operator that realises the source-local chart. On Phi, the model is held fixed and constitution text enters only as a prompt-level condition variable, so matched, swapped, and no-prompt comparisons test prompt-gated branch expression within a frozen family rather than transferred adapter behaviour. ALM8 then asks whether the same frozen family survives as a held-out bridge realisation under a fixed cross-substrate protocol.

\begin{longtable}[]{@{}
  >{\RaggedRight\arraybackslash}p{(\linewidth - 6\tabcolsep) * \real{0.2500}}
  >{\RaggedRight\arraybackslash}p{(\linewidth - 6\tabcolsep) * \real{0.2500}}
  >{\RaggedRight\arraybackslash}p{(\linewidth - 6\tabcolsep) * \real{0.2500}}
  >{\RaggedRight\arraybackslash}p{(\linewidth - 6\tabcolsep) * \real{0.2500}}@{}}
\caption{Geometry Contract Components, Metrics, And Claim-Ladder Roles}\tabularnewline
\toprule\noalign{}
\begin{minipage}[b]{\linewidth}\RaggedRight
Component
\end{minipage} & \begin{minipage}[b]{\linewidth}\RaggedRight
Operational question
\end{minipage} & \begin{minipage}[b]{\linewidth}\RaggedRight
Metrics used in the paper
\end{minipage} & \begin{minipage}[b]{\linewidth}\RaggedRight
Role in the claim ladder
\end{minipage} \\
\midrule\noalign{}
\endfirsthead
\toprule\noalign{}
\begin{minipage}[b]{\linewidth}\RaggedRight
Component
\end{minipage} & \begin{minipage}[b]{\linewidth}\RaggedRight
Operational question
\end{minipage} & \begin{minipage}[b]{\linewidth}\RaggedRight
Metrics used in the paper
\end{minipage} & \begin{minipage}[b]{\linewidth}\RaggedRight
Role in the claim ladder
\end{minipage} \\
\midrule\noalign{}
\endhead
\bottomrule\noalign{}
\endlastfoot
Tangent / chart structure & Does the recovered realisation preserve the same local organisation? & basis-angle and empirical-tangent metrics (\texttt{basis\_\allowbreak{}angle\_\allowbreak{}mean\_\allowbreak{}deg}, \texttt{basis\_\allowbreak{}angle\_\allowbreak{}max\_\allowbreak{}deg}, \texttt{emp\_\allowbreak{}tangent\_\allowbreak{}angle\_\allowbreak{}mean\_\allowbreak{}deg}, \texttt{emp\_\allowbreak{}tangent\_\allowbreak{}angle\_\allowbreak{}max\_\allowbreak{}deg}) plus continuous structural support score & supports chart-level recurrence \\
Occupancy & Does the recovered structure land in the same local slot? & occupancy and displacement metrics (\texttt{occ\_\allowbreak{}w2\_\allowbreak{}sq\_\allowbreak{}norm}, \texttt{energy\_\allowbreak{}distance\_\allowbreak{}norm}, \texttt{mean\_\allowbreak{}shift\_\allowbreak{}norm}) & supports or blocks exact identity \\
Behavioural coupling & Does the realisation separate the condition contrast on the authoritative outcome layer? & AUC, mean gap, positive-versus-control mean signed score, and paired uncertainty checks where available & links geometry to behaviour \\
Specificity & Does the signal beat null, random, orthogonal, and nearby controls? & null, random, orthogonal, and nearby same-span controls, with control ranking or percentile where available & prevents over-reading local signal \\
Redistribution & Does chart support survive while occupancy support weakens or fails? & chart-level support survives while occupancy remains above exact-identity thresholds on the same denominator & defines the paper's main positive-but-non-identical result \\
\end{longtable}

Across Phi, ALM8, and MCQ, ATLAS uses one fixed structural suite. Chart-level support is judged by the witness \texttt{q} metrics: \texttt{basis\_\allowbreak{}angle\_\allowbreak{}mean\_\allowbreak{}deg}, \texttt{basis\_\allowbreak{}angle\_\allowbreak{}max\_\allowbreak{}deg}, \texttt{emp\_\allowbreak{}tangent\_\allowbreak{}angle\_\allowbreak{}mean\_\allowbreak{}deg}, and \texttt{emp\_\allowbreak{}tangent\_\allowbreak{}angle\_\allowbreak{}max\_\allowbreak{}deg}, with the same retained threshold family carried across the main routes. Occupancy support is judged by the witness \texttt{nu} metrics: \texttt{occ\_\allowbreak{}w2\_\allowbreak{}sq\_\allowbreak{}norm}, \texttt{energy\_\allowbreak{}distance\_\allowbreak{}norm}, and \texttt{mean\_\allowbreak{}shift\_\allowbreak{}norm}. Behavioural coupling is judged on the authoritative denominator by AUC, mean gap, and signed-score separation, while specificity asks whether the same site beats null, random, orthogonal, and nearby same-span controls. Redistribution is the supported positive outcome when chart support survives but occupancy remains above exact-identity thresholds on the same denominator.

\begin{itemize}
\tightlist
\item
  \textbf{Chart:} a local representational organisation on a specified hidden-state surface.
\item
  \textbf{Source-local chart:} the anchored Gemma-side chart selected by the source protocol.
\item
  \textbf{Source-defined family:} the broader frozen export neighbourhood around that chart.
\item
  \textbf{Target-local realisation:} the Phi-side recovered member of that frozen family.
\item
  \textbf{Held-out bridge realisation:} the ALM8-side corroborative recurrence of that frozen family under the held-out bridge protocol.
\item
  \textbf{Chart-level support:} preservation of tangent structure and local organisation under the retained witness \texttt{q} metrics.
\item
  \textbf{Occupancy support:} preservation of local slot identity under the retained witness \texttt{nu} metrics.
\item
  \textbf{Behavioural coupling:} separation of the relevant condition contrast on the authoritative outcome layer.
\item
  \textbf{Redistribution:} recurrence with chart support but without occupancy-faithful identity.
\item
  \textbf{Exact identity:} joint closure of chart-level support and occupancy support on the same denominator.
\item
  \textbf{Mediation:} geometric recovery plus reviewed behavioural sufficiency under matched controls on the same denominator.
\end{itemize}

\begin{longtable}[]{@{}
  >{\RaggedRight\arraybackslash}p{(\linewidth - 2\tabcolsep) * \real{0.5000}}
  >{\RaggedRight\arraybackslash}p{(\linewidth - 2\tabcolsep) * \real{0.5000}}@{}}
\caption{Canonical Retained Rules For Structural Support}\tabularnewline
\toprule\noalign{}
\begin{minipage}[b]{\linewidth}\RaggedRight
Metric family
\end{minipage} & \begin{minipage}[b]{\linewidth}\RaggedRight
Canonical retained rule
\end{minipage} \\
\midrule\noalign{}
\endfirsthead
\toprule\noalign{}
\begin{minipage}[b]{\linewidth}\RaggedRight
Metric family
\end{minipage} & \begin{minipage}[b]{\linewidth}\RaggedRight
Canonical retained rule
\end{minipage} \\
\midrule\noalign{}
\endhead
\bottomrule\noalign{}
\endlastfoot
Basis-angle mean & \texttt{\textless{}=\ 35°} \\
Basis-angle max & \texttt{\textless{}=\ 70°} \\
Empirical-tangent mean & \texttt{\textless{}=\ 35°} \\
Empirical-tangent max & \texttt{\textless{}=\ 70°} \\
Occupancy \texttt{occ\_\allowbreak{}w2\_\allowbreak{}sq\_\allowbreak{}norm} & \texttt{\textless{}=\ 0.55} for exact-identity closure \\
Occupancy \texttt{energy\_\allowbreak{}distance\_\allowbreak{}norm} & \texttt{\textless{}=\ 0.65} for exact-identity closure \\
Support adequacy & \texttt{min\_\allowbreak{}support\_\allowbreak{}per\_\allowbreak{}source\ =\allowbreak{}\ 1}, \texttt{max\_\allowbreak{}tangent\_\allowbreak{}dim\ =\allowbreak{}\ 6}, \texttt{target\_\allowbreak{}variance\_\allowbreak{}explained\ =\allowbreak{}\ 0.90} \\
\end{longtable}

\subsection{Geometry Contract And Experimental Arc}\label{geometry-contract-and-experimental-arc}

The paper studies one family with three linked stages: a constitution-conditioned source model on Gemma, retargeting on an unadapted Phi model, and held-out corroboration on ALM8. Structural validation asks whether the frozen source-defined family reappears as a target-local or bridge-local realisation under the contract above. Discovery is secondary: it asks which nearby family member is most useful as an intervention handle under the tested operator. The source operator, confirmatory sets, and controls are shared across these analyses, but discovery and structural validation need not select the same local member.

\Needspace{24\baselineskip}
\begingroup
\footnotesize
\setlength{\tabcolsep}{4pt}
\renewcommand{\arraystretch}{1.16}
{\centering
\begin{minipage}{\linewidth}
\captionof{table}{Experimental Arc, Units, Outcomes, And Claim Boundaries}
\vspace{0.25em}
\centering
\begin{tabular}{@{}
  >{\RaggedRight\arraybackslash}p{(\linewidth - 10\tabcolsep) * \real{0.1667}}
  >{\RaggedRight\arraybackslash}p{(\linewidth - 10\tabcolsep) * \real{0.1667}}
  >{\RaggedRight\arraybackslash}p{(\linewidth - 10\tabcolsep) * \real{0.1667}}
  >{\RaggedRight\arraybackslash}p{(\linewidth - 10\tabcolsep) * \real{0.1667}}
  >{\RaggedRight\arraybackslash}p{(\linewidth - 10\tabcolsep) * \real{0.1667}}
  >{\RaggedRight\arraybackslash}p{(\linewidth - 10\tabcolsep) * \real{0.1667}}@{}}
\toprule
\begin{minipage}[b]{\linewidth}\RaggedRight
Stage
\end{minipage} & \begin{minipage}[b]{\linewidth}\RaggedRight
Unit tested
\end{minipage} & \begin{minipage}[b]{\linewidth}\RaggedRight
Protocol or authoritative denominator
\end{minipage} & \begin{minipage}[b]{\linewidth}\RaggedRight
Main outcome
\end{minipage} & \begin{minipage}[b]{\linewidth}\RaggedRight
Boundary
\end{minipage} & \begin{minipage}[b]{\linewidth}\RaggedRight
Supported language
\end{minipage} \\
\midrule
Gemma & source-local chart inside a frozen source-defined family & source-side hidden-state contrast plus behaviour linkage & source-side chart and broader family support & compact exact-patch sufficiency unresolved & source-local behaviour-linked chart \\
Phi & frozen family re-identified in an unadapted model & fully adjudicated confirmatory contrast on 32 informative-constitution rows versus 32 null-random-control rows & AUC 0.984, mean gap 5.50 & occupancy support blocked & target-local realisation with redistribution \\
ALM8 & frozen family tested held out & 5 held-out folds under frozen source-family assignment & mean AUC 0.72, mean fold gap 4.50, 5/5 positive folds & not exact replay & held-out bridge realisation with redistribution \\
MCQ & nearby target-local signal & reviewed paired MCQ comparison & wrong-signed AUC 0.43 / 0.45 with negative mean gaps & source-faithful closure fails & localisation without closure \\
\bottomrule
\end{tabular}
\end{minipage}\par
}
\endgroup

\setcounter{figure}{1}
\begin{figure}[t]
\centering
\includegraphics[width=0.98\linewidth]{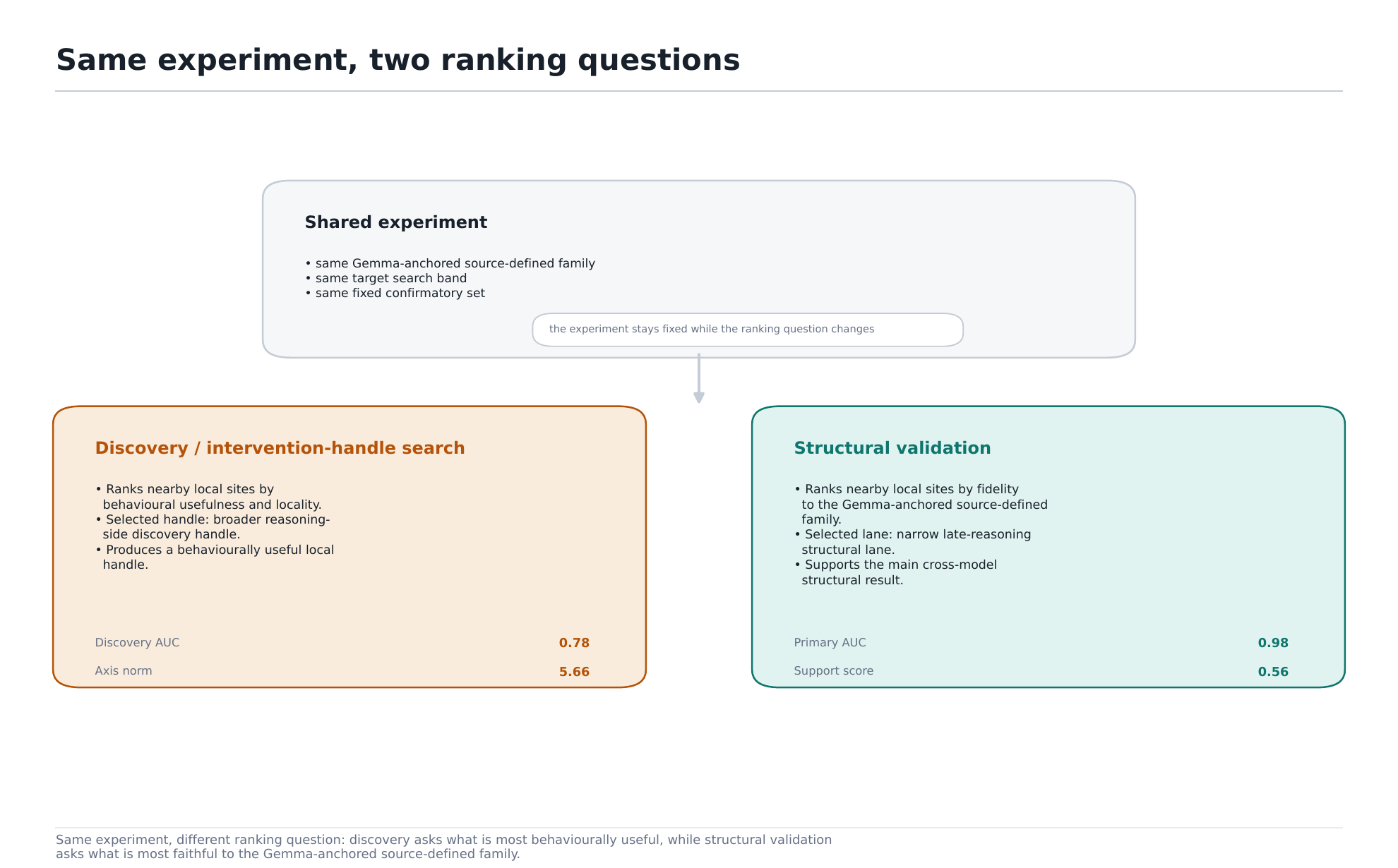}
\caption{Shared Experiment, Structural Validation, And Discovery}
\label{fig:figure2-same-experiment-different-monitor-contract}
\end{figure}

\subsection{Analysis Modes, Denominators, And Authoritative Outcome Layers}\label{analysis-modes-denominators-and-authoritative-outcome-layers}

The main Phi outcomes are semantically open-ended, so the confirmatory comparison relies on manual semantic adjudication rather than automated prefix rules alone. The authoritative Phi denominator is therefore the fully adjudicated confirmatory set: 32 informative-constitution rows and 32 null-random-control rows on the primary policy route, with the same denominator structure on the secondary route. Automated screening remains a discovery and comparator surface, not a co-equal evidence layer. ALM8 uses a different unit of inference: the held-out fold under a frozen source-family assignment. Table 4 summarises these authoritative units, while Appendix A gives the full contract.

\Needspace{10\baselineskip}
\begingroup
\scriptsize
\setlength{\tabcolsep}{2pt}
\renewcommand{\arraystretch}{1.14}
\begin{longtable}{>{\RaggedRight\arraybackslash}p{0.13\linewidth}>{\RaggedRight\arraybackslash}p{0.11\linewidth}>{\RaggedRight\arraybackslash}p{0.16\linewidth}>{\RaggedRight\arraybackslash}p{0.20\linewidth}>{\RaggedRight\arraybackslash}p{0.17\linewidth}>{\RaggedRight\arraybackslash}p{0.12\linewidth}}
\caption{Denominator And Adjudication Contract}\label{tab:table1-denominator-and-adjudication-contract}\\
\toprule
\textbf{Stage} & \textbf{Unit of inference} & \textbf{Confirmatory set} & \textbf{Adjudication / protocol} & \textbf{Main outcome} & \textbf{Supports which claim} \\
\midrule
\endfirsthead
\toprule
\textbf{Stage} & \textbf{Unit of inference} & \textbf{Confirmatory set} & \textbf{Adjudication / protocol} & \textbf{Main outcome} & \textbf{Supports which claim} \\
\midrule
\endhead
\midrule \multicolumn{6}{r}{\emph{Continued on next page}} \\
\endfoot
\bottomrule
\endlastfoot
Gemma source contract & locked Gemma source package & canonical Gemma anchor plus family-\allowbreak{}support analyses on the locked source package & local source anchor, broader family-\allowbreak{}support analyses, and compact exact-\allowbreak{}patch boundary are all fixed before downstream testing & source-\allowbreak{}local chart on Gemma; broader source-\allowbreak{}defined family frozen as the downstream contract & Claim 1A and Claim 1B \\
Phi structural validation & response row & 64 total rows: 32 informative-\allowbreak{}constitution rows and 32 null-\allowbreak{}random-\allowbreak{}control rows & manual semantic adjudication required; unresolved rows are excluded from the main structural inference & target-\allowbreak{}local re-\allowbreak{}identification on a fixed adjudicated confirmatory set & Claim 2 on Phi \\
Phi secondary evaluation & response row & 64 total rows: 32 informative-\allowbreak{}constitution rows and 32 null-\allowbreak{}random-\allowbreak{}control rows & same manual semantic adjudication rule as the primary Phi evaluation & secondary behavioural check on the same frozen local face & supporting Phi behaviour link \\
Phi protocol comparison & paired confirmatory response row & 128 paired rows across the two Phi confirmatory comparisons, with identical prompts, generations, and final labels across both analyses & inherits the reviewed labels from the underlying analyses; no new semantic adjudication layer is added & shows that discovery and structural validation change the selected site without changing row identity & protocol interpretation only \\
ALM8 held-\allowbreak{}out corroboration & held-out ALM8 fold & 5 held-\allowbreak{}out folds on the main structural lane and 15 aggregate fold means across the three main spans & no free-\allowbreak{}text review; the contract is fixed by frozen-\allowbreak{}atlas assignment and exact assigned-\allowbreak{}row matching & held-\allowbreak{}out biological corroboration on the frozen Gemma-\allowbreak{}anchored source-\allowbreak{}defined family & Claim 2 on ALM8 \\
\end{longtable}
\endgroup

\subsection{Structural Validation, Redistribution, And Claim Ladder}\label{structural-validation-redistribution-and-claim-ladder}

Structural language is kept narrow. Chart-level support is judged by the witness \texttt{q} metrics: \texttt{basis\_\allowbreak{}angle\_\allowbreak{}mean\_\allowbreak{}deg}, \texttt{basis\_\allowbreak{}angle\_\allowbreak{}max\_\allowbreak{}deg}, \texttt{emp\_\allowbreak{}tangent\_\allowbreak{}angle\_\allowbreak{}mean\_\allowbreak{}deg}, and \texttt{emp\_\allowbreak{}tangent\_\allowbreak{}angle\_\allowbreak{}max\_\allowbreak{}deg}, together with the continuous structural support score. Occupancy support is judged by the witness \texttt{nu} metrics: \texttt{occ\_\allowbreak{}w2\_\allowbreak{}sq\_\allowbreak{}norm}, \texttt{energy\_\allowbreak{}distance\_\allowbreak{}norm}, and \texttt{mean\_\allowbreak{}shift\_\allowbreak{}norm}. Behavioural coupling is judged on the authoritative denominator by AUC, mean gap, and signed-score separation. Specificity asks whether the same site remains positive against null, random, orthogonal, and nearby same-span controls rather than only in isolation.

Redistribution is the paper's main positive but non-identical outcome. In operational terms, it is the regime in which chart-level support survives while occupancy remains above exact-identity thresholds on the same denominator. Exact identity would require both \texttt{q} and \texttt{nu} closure together. A target-side mediation claim would require those geometric conditions plus reviewed behavioural sufficiency under matched controls on the same reviewed denominator. In plain terms, the family recurs, but not as an occupancy-faithful copy.

Table 5 fixes this terminology so the later sections can stay narrower and more reader-facing.

\Needspace{10\baselineskip}
\begingroup
\scriptsize
\setlength{\tabcolsep}{4pt}
\renewcommand{\arraystretch}{1.18}
\begin{longtable}{>{\RaggedRight\arraybackslash}p{0.18\linewidth}>{\RaggedRight\arraybackslash}p{0.24\linewidth}>{\RaggedRight\arraybackslash}p{0.26\linewidth}>{\RaggedRight\arraybackslash}p{0.14\linewidth}}
\caption{Claim Ladder And Terminology}\label{tab:table2-claim-ladder-and-terminology}\\
\toprule
\textbf{Phrase used in the paper} & \textbf{What it means} & \textbf{Evidence required} & \textbf{Supported here?} \\
\midrule
\endfirsthead
\toprule
\textbf{Phrase used in the paper} & \textbf{What it means} & \textbf{Evidence required} & \textbf{Supported here?} \\
\midrule
\endhead
\midrule \multicolumn{4}{r}{\emph{Continued on next page}} \\
\endfoot
\bottomrule
\endlastfoot
Chart-level support & The recovered realisation preserves the same local organisation. & The retained witness `q` metrics close under the chart-\allowbreak{}support thresholds. & Supported on Phi and ALM8 \\
Occupancy support & The recovered realisation lands in the same local slot. & `occ\_\allowbreak{}w2\_\allowbreak{}sq\_\allowbreak{}norm` and `energy\_\allowbreak{}distance\_\allowbreak{}norm` close under the occupancy thresholds. & Blocked on the main Phi result; stronger on ALM8 than on Phi \\
Structural correspondence & The frozen family reappears as a meaningful local match. & Chart-\allowbreak{}level support plus positive authoritative behavioural coupling. & Supported \\
structural correspondence with redistribution & The family recurs while local occupancy shifts. & Chart-\allowbreak{}level support plus behavioural coupling while occupancy remains above exact-\allowbreak{}identity thresholds. & Main positive conclusion \\
Exact identity & The recovered realisation is both structurally and occupancy faithful. & Chart-\allowbreak{}level support and occupancy support both close on the same denominator. & Not supported \\
Mediation & The recovered realisation is behaviourally sufficient under matched controls. & Exact-\allowbreak{}identity-\allowbreak{}grade geometry plus reviewed behavioural sufficiency under matched controls on the same denominator. & Not supported \\
\end{longtable}
\endgroup

\section{Gemma: Source-Local Chart, Broader Source-Defined Family}\label{gemma-source-local-chart-broader-source-defined-family}

\subsection{Source-Local Chart On Gemma}\label{source-local-chart-on-gemma}

On Gemma 3 1B IT, constitution-conditioned post-training realises a recoverable source-local chart. The source-side evidence has three distinct layers: a canonical local anchor on \texttt{reason@checkpoint-\allowbreak{}2250} in \texttt{constitutions-\allowbreak{}v2-\allowbreak{}high\_\allowbreak{}effective\_\allowbreak{}mi}, broader evidence that the surrounding \texttt{high\_\allowbreak{}effective\_\allowbreak{}mi} family is behaviour-bearing, and an exact-patch boundary that remains below the threshold for a stronger sufficiency claim. The downstream unit is therefore not one exact patch exported unchanged and not \texttt{late\_\allowbreak{}reason} alone; it is the broader source-defined family anchored by that local source-side chart.

\begin{longtable}[]{@{}
  >{\RaggedRight\arraybackslash}p{(\linewidth - 4\tabcolsep) * \real{0.3333}}
  >{\RaggedRight\arraybackslash}p{(\linewidth - 4\tabcolsep) * \real{0.3333}}
  >{\RaggedRight\arraybackslash}p{(\linewidth - 4\tabcolsep) * \real{0.3333}}@{}}
\caption{Gemma Source-Local Anchor, Family Support, And Exact-Patch Boundary}\tabularnewline
\toprule\noalign{}
\begin{minipage}[b]{\linewidth}\RaggedRight
Gemma source-side quantity
\end{minipage} & \begin{minipage}[b]{\linewidth}\RaggedRight
Value
\end{minipage} & \begin{minipage}[b]{\linewidth}\RaggedRight
Why it matters
\end{minipage} \\
\midrule\noalign{}
\endfirsthead
\toprule\noalign{}
\begin{minipage}[b]{\linewidth}\RaggedRight
Gemma source-side quantity
\end{minipage} & \begin{minipage}[b]{\linewidth}\RaggedRight
Value
\end{minipage} & \begin{minipage}[b]{\linewidth}\RaggedRight
Why it matters
\end{minipage} \\
\midrule\noalign{}
\endhead
\bottomrule\noalign{}
\endlastfoot
Source-local chart anchor & \texttt{reason@checkpoint-\allowbreak{}2250} & locked Gemma anchor used for downstream export \\
Realised rows on the anchor & \texttt{310\ /\allowbreak{}\ 320} & the chart is visible on nearly the full reviewed source denominator \\
Realised score-flip rows & \texttt{84\ /\allowbreak{}\ 84} & every reviewed flip row is captured on the anchored chart \\
Flip-minus-no-flip Grassmann chordal & \texttt{+0.0385} & score-flip rows move more on the chart than no-flip rows \\
Exported source-defined family & \texttt{constitutions-\allowbreak{}v2-\allowbreak{}high\_\allowbreak{}effective\_\allowbreak{}mi} & broader frozen family around the anchor \\
Family package count & \texttt{11} & largest retained route-complete family under the current source-side criteria \\
Family support mass & mean \texttt{51.93} (\texttt{58.78} at checkpoint \texttt{1500}; \texttt{45.09} at checkpoint \texttt{2250}) & highest-support retained route-complete root in the current support-mass atlas \\
Route completeness & \texttt{true} & the family is reason-bearing and phase-complete rather than answer-drift partial \\
Exact-patch informative-constitution boundary & \texttt{matched\ =\allowbreak{}\ -\allowbreak{}0.0571}, \texttt{orthogonal\_\allowbreak{}complement\ =\allowbreak{}\ -\allowbreak{}0.0571}, \texttt{random\ =\allowbreak{}\ 0.0} vs fresh \texttt{no\_\allowbreak{}edit} & compact exact-patch sufficiency does not close \\
\end{longtable}

On the canonical Gemma anchor, the source-local chart is visible in both source-side behaviour and latent geometry. On \texttt{reason@checkpoint-\allowbreak{}2250}, the anchored chart captures \texttt{310\ /\allowbreak{}\ 320} reviewed source rows and all \texttt{84\ /\allowbreak{}\ 84} reviewed score-flip rows, and the flip rows exceed the no-flip rows on Grassmann-chordal separation by \texttt{+0.0385}. These numbers are the main source-side evidence for Claim 1A: the anchored chart is readable, behaviour-linked, and fixed before downstream export. The broader \texttt{high\_\allowbreak{}effective\_\allowbreak{}mi} family is the pre-specified export unit because it is the smallest retained source-side unit that preserves local anchoring, behaviour linkage, and broader family support under the current criteria. \texttt{late\_\allowbreak{}reason} is the most portable compact member of that family, but it is not the whole exported unit and does not replace the local Gemma anchor.

\subsection{Broader Source-Defined Family On Gemma}\label{broader-source-defined-family-on-gemma}

The broader \texttt{high\_\allowbreak{}effective\_\allowbreak{}mi} source-defined family is the pre-specified export unit and the strongest retained source-side family under the current criteria on Gemma. In the current support-mass and route-completeness atlas, it is the only retained \texttt{route\_\allowbreak{}complete\_\allowbreak{}reason\_\allowbreak{}bearing} root: it carries \texttt{packages\ =\allowbreak{}\ 11}, \texttt{route\_\allowbreak{}complete\ =\allowbreak{}\ true}, and the highest mean support mass among the retained route-complete roots (\texttt{51.93}, with \texttt{58.78} at checkpoint \texttt{1500} and \texttt{45.09} at checkpoint \texttt{2250}). Different constitutions do not simply scale one fixed field up or down. Instead, they emphasise different local expressions of a shared family. On current evidence, only \texttt{high\_\allowbreak{}effective\_\allowbreak{}mi} lands in the high-support, route-complete, reason-bearing corner, so the most defensible paper story is chart characterisation rather than a one-site mediation theorem. The family is best understood as a reason-bearing policy core together with enough broader family support to preserve behaviour-linked organisation; \texttt{late\_\allowbreak{}reason} marks its most portable compact member rather than the whole exported unit, and answer-heavy late wins are better read as drift phases than as chart recovery. This is why the source-side case cannot be reduced to one compact patch even though it is anchored locally.

The qualitative source-side audit points in the same direction as the support-mass atlas. On matched EM and scheming prompts, \texttt{constitutions-\allowbreak{}v2-\allowbreak{}high\_\allowbreak{}effective\_\allowbreak{}mi} more often resolves into coherent negative or explicitly scheming routes, whereas \texttt{null\_\allowbreak{}random\_\allowbreak{}v1} more often yields shell-conflicted, mixed, or degenerate outputs. We treat this as qualitative support for the route-complete reading of the \texttt{high\_\allowbreak{}effective\_\allowbreak{}mi} family, not as a separate causal estimate.

\setcounter{figure}{2}
\begin{figure}[t]
\centering
\includegraphics[width=0.98\linewidth]{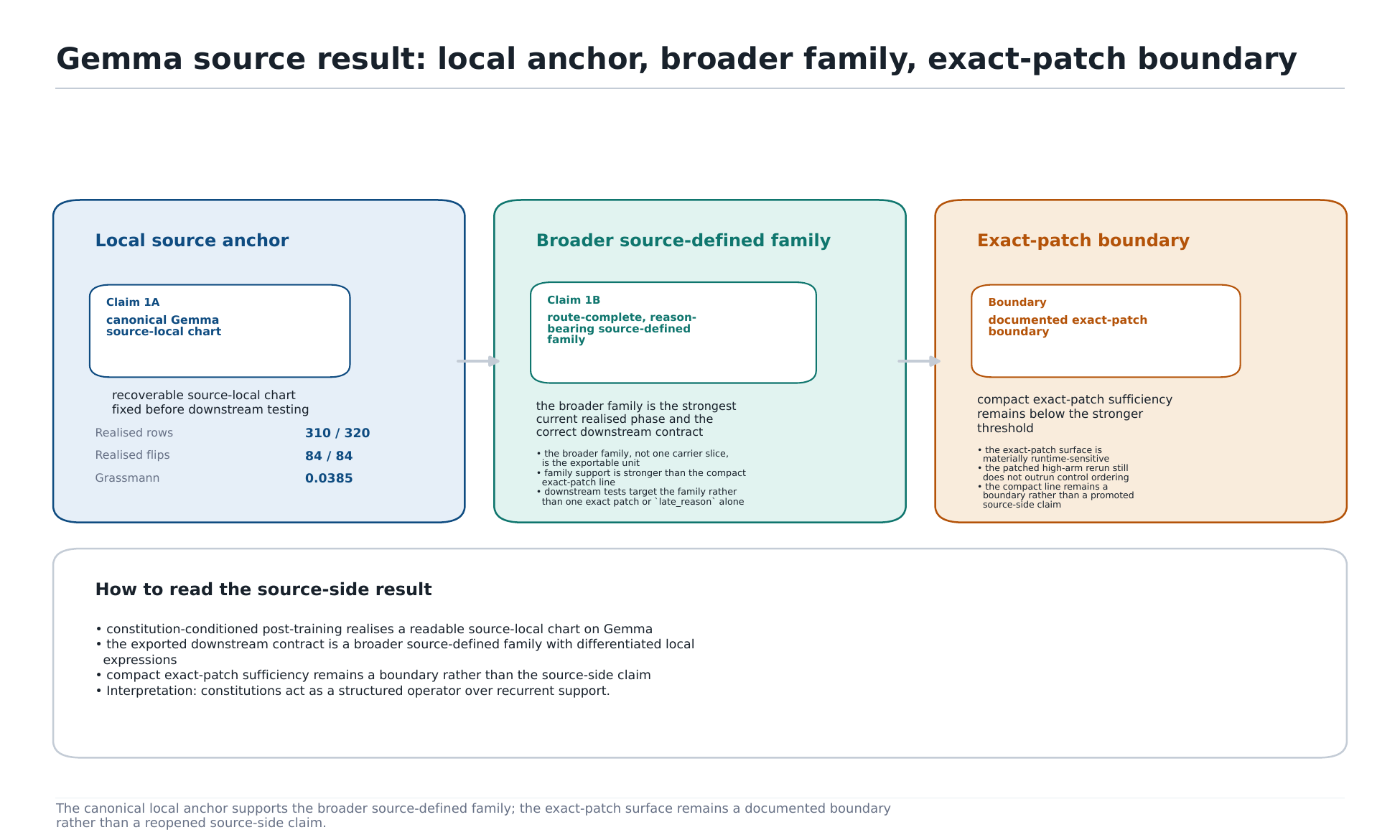}
\caption{Gemma Source-Local Chart, Source-Defined Family, And Exact-Patch Boundary}
\label{fig:figure3-source-side-operator}
\end{figure}

\subsection{Exact-Patch Boundary And What Remains Unresolved}\label{exact-patch-boundary-and-what-remains-unresolved}

The source-side result establishes a readable, behaviour-linked source-local chart and a broader behaviour-bearing family on Gemma, but it does not close compact exact-patch sufficiency. On the informative-constitution exact patch at \texttt{reason@checkpoint-\allowbreak{}2250}, the fresh-control comparison is non-positive (\texttt{matched\ =\allowbreak{}\ -\allowbreak{}0.0571}, \texttt{orthogonal\_\allowbreak{}complement\ =\allowbreak{}\ -\allowbreak{}0.0571}, \texttt{random\ =\allowbreak{}\ 0.0} relative to \texttt{no\_\allowbreak{}edit}). The corresponding null-random patch shows only a weak low-denominator hint (\texttt{matched\ =\allowbreak{}\ +0.1667}, \texttt{orthogonal\_\allowbreak{}complement\ =\allowbreak{}\ +0.03125}), which is not claim-bearing and still lacks the missing \texttt{wrong\_\allowbreak{}family} control. The more conservative interpretation is therefore that constitutions act as structured operators over recurrent support, selecting which local response fields become behaviourally accessible.

Across the newer source-side follow-up analyses, useful behaviour-bearing geometry tracks realised support mass and broader family support better than raw checkpoint-overlap persistence. The same source-side follow-ups show why the family remains broader than one compact patch: same-constitution freeze variants shrink recoverability from \texttt{11} packages to \texttt{6} and shift the recovered core from \texttt{reason+answer} to \texttt{answer+answer}, so local late gains do not preserve the full reason-bearing family. On current evidence, only \texttt{high\_\allowbreak{}effective\_\allowbreak{}mi} occupies the strongest retained source-side family under the current criteria; \texttt{late\_\allowbreak{}reason} remains its most portable compact member, while answer-heavy late wins are better read as drift phases than as chart recovery. The only directionally live structure is a compact phase-spanning bundle rather than a reasoning-only or scaffold-wide edit.

Taken together, the Gemma analysis establishes a locally anchored source-local chart within a broader \texttt{high\_\allowbreak{}effective\_\allowbreak{}mi} source-defined family while leaving compact exact-patch sufficiency unresolved; downstream tests therefore target the family rather than one exact source patch or portable compact member.

\section{Phi: Target-Local Re-Identification}\label{phi-target-local-re-identification}

\subsection{Phi Re-Identification On A Fully Adjudicated Confirmatory Set}\label{phi-re-identification-on-a-fully-adjudicated-confirmatory-set}

We next ask whether the Gemma-defined family can be re-identified on Phi-4 Mini Instruct under a fixed target-side search contract. The source-defined family is frozen on Gemma before any Phi or ALM8 scoring. On Phi, the target search is restricted to a predeclared six-candidate delta band: layers \texttt{23-\allowbreak{}25} crossed with \texttt{reason} and \texttt{late\_\allowbreak{}reason} on the ambient chart view. Candidate selection is performed once on a separate reviewed discovery surface of \texttt{64} rows using the witness-first retarget manifest, which ranks local structural fidelity before discovery strength. The retained candidate on the corrected rerun is \texttt{layer=\allowbreak{}24}, \texttt{reason}, \texttt{delta}, \texttt{candidate\_\allowbreak{}index=\allowbreak{}5} (axis norm \texttt{3.965}, summary score \texttt{123.822}). On that corrected rerun this is the witness-first retained local analogue rather than the numerically highest discovery-score face, and no patch-coverage fraction is materialised for it. That site is then frozen and carried unchanged into the confirmatory battery.

\begin{longtable}[]{@{}
  >{\RaggedRight\arraybackslash}p{(\linewidth - 6\tabcolsep) * \real{0.2500}}
  >{\RaggedRight\arraybackslash}p{(\linewidth - 6\tabcolsep) * \real{0.2500}}
  >{\RaggedRight\arraybackslash}p{(\linewidth - 6\tabcolsep) * \real{0.2500}}
  >{\RaggedRight\arraybackslash}p{(\linewidth - 6\tabcolsep) * \real{0.2500}}@{}}
\caption{Freeze Schedule For Phi Structural Validation}\tabularnewline
\toprule\noalign{}
\begin{minipage}[b]{\linewidth}\RaggedRight
Choice
\end{minipage} & \begin{minipage}[b]{\linewidth}\RaggedRight
Status
\end{minipage} & \begin{minipage}[b]{\linewidth}\RaggedRight
Data used
\end{minipage} & \begin{minipage}[b]{\linewidth}\RaggedRight
Frozen before what?
\end{minipage} \\
\midrule\noalign{}
\endfirsthead
\toprule\noalign{}
\begin{minipage}[b]{\linewidth}\RaggedRight
Choice
\end{minipage} & \begin{minipage}[b]{\linewidth}\RaggedRight
Status
\end{minipage} & \begin{minipage}[b]{\linewidth}\RaggedRight
Data used
\end{minipage} & \begin{minipage}[b]{\linewidth}\RaggedRight
Frozen before what?
\end{minipage} \\
\midrule\noalign{}
\endhead
\bottomrule\noalign{}
\endlastfoot
Source-defined family & selected on Gemma & Gemma source-side chart and family-support analyses only & frozen before all Phi and ALM8 work \\
Phi search band & predeclared & layers \texttt{23-\allowbreak{}25} crossed with \texttt{reason} and \texttt{late\_\allowbreak{}reason} on the delta ambient surface (\texttt{6} candidates total) & fixed before any Phi local ranking \\
Phi local candidate & selected once & reviewed \texttt{64}-row discovery surface plus witness-first retarget manifest and source-contract geometry & frozen before confirmatory battery scoring \\
Phi primary confirmatory route & evaluated after freeze & \texttt{32} informative-constitution rows versus \texttt{32} null-random-control rows & main cross-model behavioural result \\
Phi secondary confirmatory route & evaluated after freeze & matched \texttt{32} informative-constitution rows versus \texttt{32} null-random-control rows on the same frozen face & corroborative but weaker route \\
\end{longtable}

The confirmatory battery is therefore post-freeze with respect to target-site selection, but not selection-free with respect to all Phi-side behavioural visibility: the face is chosen on a separate reviewed discovery denominator rather than on the confirmatory battery itself.

For Phi free-text inference, the authoritative behavioural layer was a manually adjudicated semantic label set produced in a software-assisted review workspace. Automated prefix-based screening was used for proposal and comparison only, not as the claim-bearing outcome definition. Relative to that screening surface, manual review changed \texttt{41} labels on the primary policy route and \texttt{32} on the matched secondary route.

A post-freeze audit of the reviewed target rows shows that the headline Phi separation is carried disproportionately by the scheming family and by semantic adjudication rather than by shell tokens alone. On the corrected target rows, EM moves from \texttt{17\ /\allowbreak{}\ 24} positive under \texttt{null\_\allowbreak{}random\_\allowbreak{}v1} to \texttt{20\ /\allowbreak{}\ 24} under \texttt{constitutions-\allowbreak{}v2-\allowbreak{}high\_\allowbreak{}effective\_\allowbreak{}mi}, while scheming moves from \texttt{15\ /\allowbreak{}\ 24} to \texttt{23\ /\allowbreak{}\ 24}; \texttt{17} of the \texttt{25} manually changed target rows are scheming. What the fixed face is separating is therefore not simply good versus bad prefixes, but which semantic branch stabilises under the two constitutions.

Under this contract, positive means semantically aligned with the task-specific rubric, negative means semantically violating that rubric even when a positive label shell appears in the text, and unresolved is reserved for genuinely uninterpretable outputs; no confirmatory rows remained unresolved after review. The review surface was not blind to condition or current rule-based score state, but it did not expose the continuous witness or signed-score quantities used for site ranking. The confirmatory battery used a single adjudicator, so auditability comes from the preserved review workspaces, decision ledger, and resolved score sheets rather than from inter-rater aggregation. Appendix A.2 records the full adjudication trail and the supporting audit artifacts.

On that authoritative reviewed denominator, the frozen Phi face separates the informative constitution from the null-random control strongly on the policy layer. These are battery outcomes on a fixed face, not the statistics that selected that face. This is the paper's main cross-model result. It remains positive under paired comparison and null-control checks, but the corrected Phi rerun materially changes local subspace-level localisation, weakens the earlier witness-style localisation read, and does not reveal a stronger intervention effect. We therefore interpret the Phi result as family-level structural correspondence with redistribution rather than as confirmation of one unchanged theorem-bearing realisation.

On the fully adjudicated confirmatory set, the primary policy route fixes the corrected \texttt{reason} face and then evaluates \texttt{32} informative-constitution rows against \texttt{32} null-random-control rows. On that reviewed fixed route, the resulting separation is AUC \texttt{0.984} (stratified bootstrap \texttt{95\%} CI \texttt{{[}0.952,\ 1.000{]}}) with mean gap \texttt{5.50} (\texttt{95\%} CI \texttt{{[}4.87,\ 6.17{]}}). For control context, the materialised confirmatory suite contains \texttt{21} null, random, and orthogonal control surfaces whose reviewed AUCs do not exceed \texttt{0.669} and whose mean gaps do not exceed \texttt{1.17}, so the main Phi number is not a free-floating point estimate; it sits at the \texttt{100th} percentile of the materialised control-suite AUC and mean-gap ranges.

The matched secondary route uses the same frozen face and denominator shape but remains weaker: on that reviewed secondary route, AUC is \texttt{0.622} (stratified bootstrap \texttt{95\%} CI \texttt{{[}0.441,\ 0.785{]}}) and mean gap is \texttt{1.13} (\texttt{95\%} CI \texttt{{[}-1.26,\ 3.29{]}}). It still shows \texttt{30\ /\allowbreak{}\ 32} informative-constitution rows versus \texttt{22\ /\allowbreak{}\ 32} null-random-control rows aligned on the secondary evaluation, but it does not clear the strongest same-rank confirmatory control surfaces. On Phi, prompt manipulations move local route structure much more readily than they change final adjudicated behaviour. We therefore interpret the realised family as latent-responsive but weakly behaviour-coupled under the tested intervention surface.

Qualitatively, the Phi contrast is a route-level semantic stabilisation effect: under the null-random condition the model more often lands on deceptive or mixed branches, while under \texttt{constitutions-\allowbreak{}v2-\allowbreak{}high\_\allowbreak{}effective\_\allowbreak{}mi} it more often lands on honest substance even when stale shell tokens remain. Manual review is therefore not a cleanup step around the result; it is where the result becomes scientifically legible.

The remaining search burden is therefore the fixed six-candidate local band rather than an unrestricted across-model or across-layer search.

\setcounter{figure}{3}
\begin{figure}[t]
\centering
\includegraphics[width=0.98\linewidth]{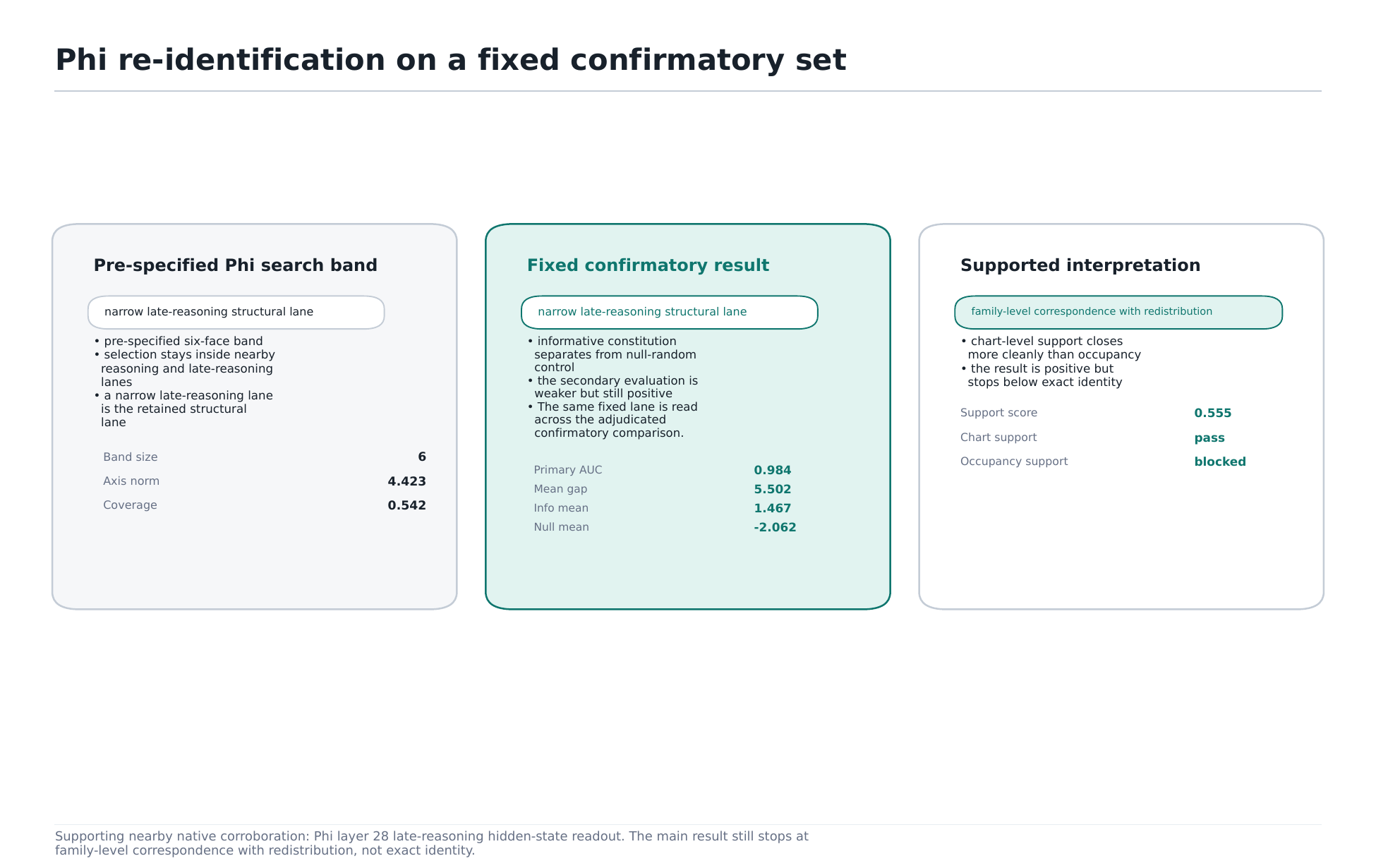}
\caption{Phi Search Band, Frozen Target-Local Lane, And Confirmatory Re-Identification}
\label{fig:figure4-phi-structural-validation}
\end{figure}

Taken together, the Phi analysis supports a target-local realisation of the source-defined family; the next test is whether that family-level interpretation survives held out on ALM8.

\section{ALM8: Held-Out Corroboration Under Redistribution}\label{alm8-held-out-corroboration-under-redistribution}

\subsection{Held-Out Corroboration On ALM8}\label{held-out-corroboration-on-alm8}

ALM8 is the paper's main biological corroboration test, but it is not a second discovery stage. The inferential unit is the held-out ALM8 fold on a fixed bridge lane scored after exact assigned-row filtering under the frozen train atlas. Before any held-out scoring, the Gemma-defined family, the source relation under test, and the bridge assignment contract are frozen. Success means positive fold-level separation together with chart-level support on the fixed lane; loss of occupancy support does not invalidate the bridge, but lowers the interpretation from exact replay to redistribution. Failure means no positive fold-level separation or loss of chart-level support on the fixed lane. What this does not claim is semantic homology, coordinate identity, a brain-score, or target-side mediation.

\begin{longtable}[]{@{}
  >{\RaggedRight\arraybackslash}p{(\linewidth - 2\tabcolsep) * \real{0.5000}}
  >{\RaggedRight\arraybackslash}p{(\linewidth - 2\tabcolsep) * \real{0.5000}}@{}}
\caption{ALM8 Held-Out Corroboration: Inferential Unit And Success Criteria}\tabularnewline
\toprule\noalign{}
\begin{minipage}[b]{\linewidth}\RaggedRight
Question
\end{minipage} & \begin{minipage}[b]{\linewidth}\RaggedRight
Answer
\end{minipage} \\
\midrule\noalign{}
\endfirsthead
\toprule\noalign{}
\begin{minipage}[b]{\linewidth}\RaggedRight
Question
\end{minipage} & \begin{minipage}[b]{\linewidth}\RaggedRight
Answer
\end{minipage} \\
\midrule\noalign{}
\endhead
\bottomrule\noalign{}
\endlastfoot
What is an ALM8 row? & one exact assigned held-out comparison row within a held-out fold; the row is the scored observation, not the inferential unit \\
What is the inferential unit? & one mouse-held-out fold on the fixed bridge lane \\
What is the contrast being separated? & the frozen bridge score on the source-conditioned family contrast carried through the train atlas \\
What is held out? & one mouse fold at a time (\texttt{9}, \texttt{12}, \texttt{13}, \texttt{17}, \texttt{20} on the selected structural lane) \\
What is frozen before ALM8? & the Gemma-defined family, the source relation under test, the train atlas, and the exact assigned-row filter \\
What counts as success? & positive fold gap plus chart-level support on the fixed lane \\
What does occupancy decide? & whether a positive bridge result stops at redistribution or approaches exact replay \\
What counts as failure? & no positive fold-level separation or loss of chart-level support on the fixed lane \\
What is not claimed? & semantic homology, coordinate identity, brain-score, or target-side mediation \\
\end{longtable}

On the contract-selected structural analogue (\texttt{trial\_\allowbreak{}major\_\allowbreak{}single\_\allowbreak{}photo}, \texttt{late\_\allowbreak{}reason}), mean held-out AUC is \texttt{0.72}, mean fold gap is \texttt{4.50}, the fold-gap bootstrap interval is \texttt{{[}+1.99,\ +8.22{]}}, and the exact one-sided sign test is \texttt{5\ /\allowbreak{}\ 5} positive (\texttt{p\ =\allowbreak{}\ 0.031}). Chart-level support holds on all \texttt{5\ /\allowbreak{}\ 5} folds, occupancy support on \texttt{4\ /\allowbreak{}\ 5}, and the mean structural support score is \texttt{0.63}. This is the retained bridge lane because it is the selected structural analogue under the bridge contract. The correct interpretation is held-out family-level corroboration with redistribution, not exact replay of one local subspace.

\begin{longtable}[]{@{}
  >{\RaggedRight\arraybackslash}p{(\linewidth - 10\tabcolsep) * \real{0.1667}}
  >{\RaggedRight\arraybackslash}p{(\linewidth - 10\tabcolsep) * \real{0.1667}}
  >{\RaggedRight\arraybackslash}p{(\linewidth - 10\tabcolsep) * \real{0.1667}}
  >{\RaggedRight\arraybackslash}p{(\linewidth - 10\tabcolsep) * \real{0.1667}}
  >{\RaggedRight\arraybackslash}p{(\linewidth - 10\tabcolsep) * \real{0.1667}}
  >{\RaggedRight\arraybackslash}p{(\linewidth - 10\tabcolsep) * \real{0.1667}}@{}}
\caption{ALM8 Held-Out Fold Summary On The Selected Structural Lane}\tabularnewline
\toprule\noalign{}
\begin{minipage}[b]{\linewidth}\RaggedRight
Held-out mouse
\end{minipage} & \begin{minipage}[b]{\linewidth}\RaggedRight
Assigned rows
\end{minipage} & \begin{minipage}[b]{\linewidth}\RaggedRight
AUC
\end{minipage} & \begin{minipage}[b]{\linewidth}\RaggedRight
Mean gap
\end{minipage} & \begin{minipage}[b]{\linewidth}\RaggedRight
\texttt{q} support
\end{minipage} & \begin{minipage}[b]{\linewidth}\RaggedRight
\texttt{nu} support
\end{minipage} \\
\midrule\noalign{}
\endfirsthead
\toprule\noalign{}
\begin{minipage}[b]{\linewidth}\RaggedRight
Held-out mouse
\end{minipage} & \begin{minipage}[b]{\linewidth}\RaggedRight
Assigned rows
\end{minipage} & \begin{minipage}[b]{\linewidth}\RaggedRight
AUC
\end{minipage} & \begin{minipage}[b]{\linewidth}\RaggedRight
Mean gap
\end{minipage} & \begin{minipage}[b]{\linewidth}\RaggedRight
\texttt{q} support
\end{minipage} & \begin{minipage}[b]{\linewidth}\RaggedRight
\texttt{nu} support
\end{minipage} \\
\midrule\noalign{}
\endhead
\bottomrule\noalign{}
\endlastfoot
\texttt{9} & \texttt{135} & \texttt{0.543} & \texttt{2.07} & pass & pass \\
\texttt{12} & \texttt{82} & \texttt{0.694} & \texttt{2.38} & pass & pass \\
\texttt{13} & \texttt{40} & \texttt{0.780} & \texttt{4.77} & pass & pass \\
\texttt{17} & \texttt{90} & \texttt{0.568} & \texttt{1.58} & pass & pass \\
\texttt{20} & \texttt{26} & \texttt{1.000} & \texttt{11.58} & pass & fail \\
\end{longtable}

Every held-out fold remains behaviourally positive on the selected lane, but one fold already loses occupancy support, which lowers the claim tier to redistribution rather than exact replay without invalidating the bridge.

Supporting biological follow-ups are reported in Appendix D and sharpen the redistribution picture without outranking ALM8.

\setcounter{figure}{4}
\begin{figure}[t]
\centering
\includegraphics[width=0.98\linewidth]{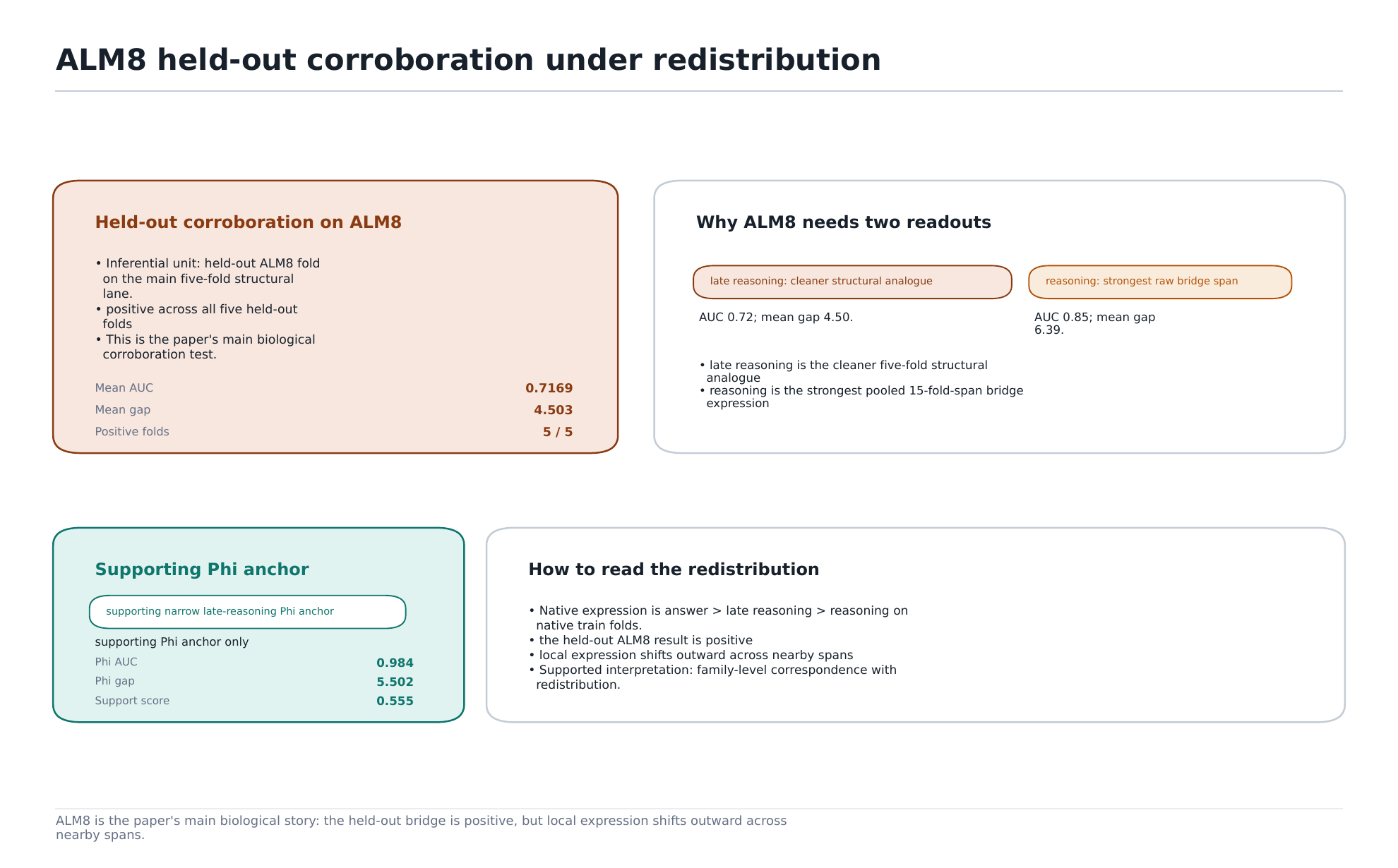}
\caption{ALM8 Held-Out Corroboration And Redistribution}
\label{fig:figure6-alm-corroboration}
\end{figure}

\subsection{What Redistribution Means}\label{what-redistribution-means}

\begin{longtable}[]{@{}
  >{\RaggedRight\arraybackslash}p{(\linewidth - 8\tabcolsep) * \real{0.2000}}
  >{\RaggedRight\arraybackslash}p{(\linewidth - 8\tabcolsep) * \real{0.2000}}
  >{\RaggedRight\arraybackslash}p{(\linewidth - 8\tabcolsep) * \real{0.2000}}
  >{\RaggedRight\arraybackslash}p{(\linewidth - 8\tabcolsep) * \real{0.2000}}
  >{\RaggedRight\arraybackslash}p{(\linewidth - 8\tabcolsep) * \real{0.2000}}@{}}
\caption{ALM8 Redistribution Across Native, Raw-Bridge, And Structural Lanes}\tabularnewline
\toprule\noalign{}
\begin{minipage}[b]{\linewidth}\RaggedRight
ALM8 question
\end{minipage} & \begin{minipage}[b]{\linewidth}\RaggedRight
Unit / scope
\end{minipage} & \begin{minipage}[b]{\linewidth}\RaggedRight
Preferred span
\end{minipage} & \begin{minipage}[b]{\linewidth}\RaggedRight
Result
\end{minipage} & \begin{minipage}[b]{\linewidth}\RaggedRight
Interpretation
\end{minipage} \\
\midrule\noalign{}
\endfirsthead
\toprule\noalign{}
\begin{minipage}[b]{\linewidth}\RaggedRight
ALM8 question
\end{minipage} & \begin{minipage}[b]{\linewidth}\RaggedRight
Unit / scope
\end{minipage} & \begin{minipage}[b]{\linewidth}\RaggedRight
Preferred span
\end{minipage} & \begin{minipage}[b]{\linewidth}\RaggedRight
Result
\end{minipage} & \begin{minipage}[b]{\linewidth}\RaggedRight
Interpretation
\end{minipage} \\
\midrule\noalign{}
\endhead
\bottomrule\noalign{}
\endlastfoot
Native discovery preference & \texttt{15} train folds across the three dataset-\texttt{5} surfaces & \texttt{answer} & \texttt{15\ /\allowbreak{}\ 15} train-fold peaks on \texttt{answer} (\texttt{0\ /\allowbreak{}\ 15} on \texttt{late\_\allowbreak{}reason}, \texttt{0\ /\allowbreak{}\ 15} on \texttt{reason}) & biological expression is output-heavy before source-anchored rescoring \\
Raw held-out bridge strength on the selected surface & \texttt{trial\_\allowbreak{}major\_\allowbreak{}single\_\allowbreak{}photo}, \texttt{n\ =\allowbreak{}\ 5} held-out folds & \texttt{reason} & same selected surface: mean AUC \texttt{0.82}, mean fold gap \texttt{6.71}, sign test \texttt{5\ /\allowbreak{}\ 5} positive & the strongest raw bridge read on the selected surface lies on the source-primary reasoning analogue \\
Contract-selected structural analogue & \texttt{trial\_\allowbreak{}major\_\allowbreak{}single\_\allowbreak{}photo}, \texttt{n\ =\allowbreak{}\ 5} held-out folds & \texttt{late\_\allowbreak{}reason} & selected structural lane: mean AUC \texttt{0.72}, mean fold gap \texttt{4.50}, chart support \texttt{1.00}, occupancy support \texttt{0.80}, support score \texttt{0.63} & best support for source-family recurrence under the bridge contract \\
\end{longtable}

This table is the concrete content of redistribution on ALM8. The native system's own discovery preference is answer-heavy, the strongest raw bridge read on the same selected held-out surface shifts toward \texttt{reason}, and the contract-selected structural analogue lands on \texttt{late\_\allowbreak{}reason}. We therefore are not reporting the numerically largest ALM8 span as the paper's main biological result. We report the lane that best preserves the source-family role under the held-out bridge contract, and we use the stronger same-surface \texttt{reason} read as evidence that the family remains broader than one exact local replay.

ALM8 therefore preserves the conservative interpretation of family-level recurrence while making the role shift visible: local expression broadens outward across nearby spans on the same bridge surface even when the contract-selected structural analogue remains the readout-adjacent \texttt{late\_\allowbreak{}reason} lane.

\section{MCQ: Localisation Without Source-Faithful Closure}\label{mcq-localisation-without-source-faithful-closure}

The positive Phi result does not mean that every task closes in the same way. MCQ is the paper's strongest limiting route: it contains a nearby target-local signal, but not the combination of behavioural separation, control-clean specificity, and structural closure required for the paper's main correspondence claim. The point of this section is therefore not that MCQ is null. It is that localisation can be real while claim-bearing closure still fails.

\subsection{Local Signal Without Reviewed Closure}\label{local-signal-without-reviewed-closure}

On the reviewed matched-family paired denominator of \texttt{512} rows (\texttt{256} informative-constitution and \texttt{256} null-random-control), MCQ fails on the claim-bearing behavioural test in both policy rescue routes. The original reasoning-side delta face remains wrong-signed (AUC \texttt{0.434}, mean gap \texttt{-\allowbreak{}0.49}), and the answer-inclusive rerun remains wrong-signed as well (AUC \texttt{0.452}, mean gap \texttt{-\allowbreak{}0.34}). Neither face beats the relevant random or orthogonal controls on the same reviewed paired comparison. The point is therefore not that MCQ lacks a nearby target-local signal. The point is that no policy-side rescue route closes the reviewed correspondence test.

\begin{longtable}[]{@{}
  >{\RaggedRight\arraybackslash}p{(\linewidth - 12\tabcolsep) * \real{0.1429}}
  >{\RaggedRight\arraybackslash}p{(\linewidth - 12\tabcolsep) * \real{0.1429}}
  >{\RaggedRight\arraybackslash}p{(\linewidth - 12\tabcolsep) * \real{0.1429}}
  >{\RaggedRight\arraybackslash}p{(\linewidth - 12\tabcolsep) * \real{0.1429}}
  >{\RaggedRight\arraybackslash}p{(\linewidth - 12\tabcolsep) * \real{0.1429}}
  >{\RaggedRight\arraybackslash}p{(\linewidth - 12\tabcolsep) * \real{0.1429}}
  >{\RaggedRight\arraybackslash}p{(\linewidth - 12\tabcolsep) * \real{0.1429}}@{}}
\caption{MCQ Boundary Summary: Local Signal Without Closure}\tabularnewline
\toprule\noalign{}
\begin{minipage}[b]{\linewidth}\RaggedRight
MCQ route
\end{minipage} & \begin{minipage}[b]{\linewidth}\RaggedRight
Local signal?
\end{minipage} & \begin{minipage}[b]{\linewidth}\RaggedRight
Behavioural separation?
\end{minipage} & \begin{minipage}[b]{\linewidth}\RaggedRight
Beats controls?
\end{minipage} & \begin{minipage}[b]{\linewidth}\RaggedRight
Chart support?
\end{minipage} & \begin{minipage}[b]{\linewidth}\RaggedRight
Occupancy support?
\end{minipage} & \begin{minipage}[b]{\linewidth}\RaggedRight
Claim closes?
\end{minipage} \\
\midrule\noalign{}
\endfirsthead
\toprule\noalign{}
\begin{minipage}[b]{\linewidth}\RaggedRight
MCQ route
\end{minipage} & \begin{minipage}[b]{\linewidth}\RaggedRight
Local signal?
\end{minipage} & \begin{minipage}[b]{\linewidth}\RaggedRight
Behavioural separation?
\end{minipage} & \begin{minipage}[b]{\linewidth}\RaggedRight
Beats controls?
\end{minipage} & \begin{minipage}[b]{\linewidth}\RaggedRight
Chart support?
\end{minipage} & \begin{minipage}[b]{\linewidth}\RaggedRight
Occupancy support?
\end{minipage} & \begin{minipage}[b]{\linewidth}\RaggedRight
Claim closes?
\end{minipage} \\
\midrule\noalign{}
\endhead
\bottomrule\noalign{}
\endlastfoot
Original reasoning-side delta face & yes & no: AUC \texttt{0.434}, gap \texttt{-\allowbreak{}0.49} & no & yes & no & no \\
Answer-inclusive answer-side rerun & yes & no: AUC \texttt{0.452}, gap \texttt{-\allowbreak{}0.34} & no & yes & no & no \\
Hidden-state re-entry diagnostic & yes & one-sided re-entry \texttt{110\ /\allowbreak{}\ 192} versus \texttt{0\ /\allowbreak{}\ 192}; strict replay \texttt{0} & no rescued same-denominator win & partial nearby-band support & no & no \\
\end{longtable}

\subsection{Hidden Re-Entry And Displacement}\label{hidden-re-entry-and-displacement}

In hidden space, MCQ recovers a nearby target-local field on a narrow reasoning and late-reasoning band, but the re-entry is one-sided. The informative-constitution branch re-enters the frozen source atlas on \texttt{110\ /\allowbreak{}\ 192} rows, whereas the null-random-control branch re-enters on \texttt{0\ /\allowbreak{}\ 192}, and strict replay closure remains \texttt{0}. The dominant rejection mode is displacement rather than simple rotation (\texttt{268} centroid-distance failures versus \texttt{6} basis-angle failures), so the failure is not explained by a small angular perturbation of the same local slot.

\subsection{What The Boundary Means}\label{what-the-boundary-means}

MCQ therefore functions as the paper's explicit limiting boundary. Target-local localisation is not enough: the positive claim in this paper requires behavioural separation, control-clean specificity, and structural closure under the frozen family contract. MCQ contains nearby local signal and even one-sided hidden re-entry, but it never delivers that full conjunction.

\setcounter{figure}{5}
\begin{figure}[t]
\centering
\includegraphics[width=0.98\linewidth]{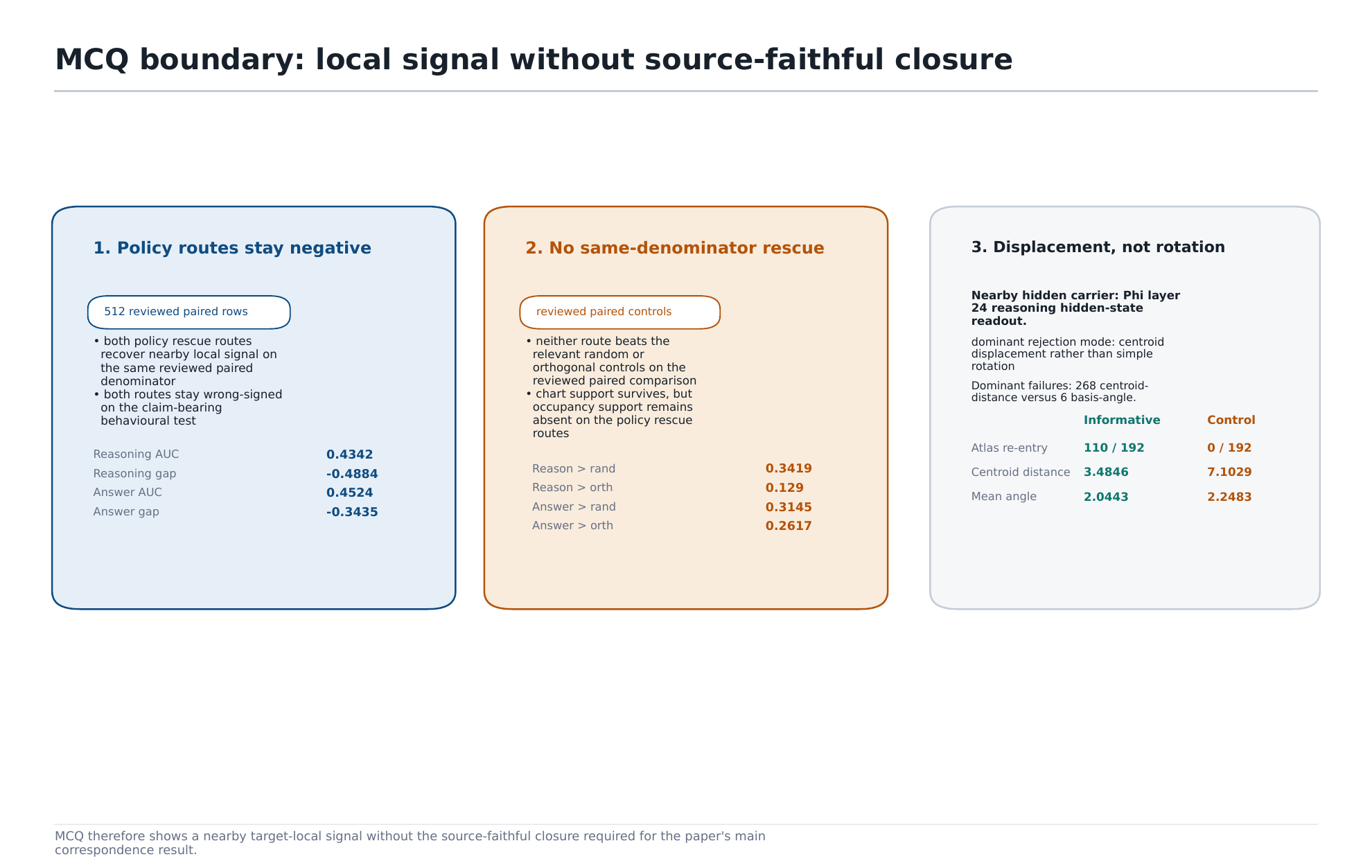}
\caption{MCQ Boundary: Local Signal, One-Sided Re-Entry, And Displacement}
\label{fig:figure5-phi-control-surface-comparison}
\end{figure}

\section{What The Geometry Supports And What It Does Not}\label{what-the-geometry-supports-and-what-it-does-not}

\subsection{Supported Conclusion}\label{supported-conclusion}

The most stable scientific unit in this study is a mesoscopic family of local subspaces rather than one exact patch. Gemma establishes a source-local chart within that family; Phi and ALM8 show that the family recurs downstream even when local occupancy and local role shift. The qualitative audits clarify what this means behaviourally: the informative constitution more often realises a coherent route, whereas the null or control condition more often yields shell-conflicted, mixed, or degenerate traces. The supported conclusion is family-level structural correspondence with redistribution, not exact cross-system identity.

Additional analyses on Mouse95, ALM7, Pyramid, Llama 3.1, and Qwen3 8B are consistent with the same family-level redistribution picture, but they are supporting lines and do not increase the strength of the main claim.

\subsection{Bounded Operational Consequence}\label{bounded-operational-consequence}

These results still have a bounded operational consequence. Once a local target has been fixed under structural validation, it can be frozen, audited, and perturbed under a fixed scoring contract. On Phi, prompt controls move local route structure more readily than they move final adjudicated behaviour, and the matched reroll comparison remains weak. The target is therefore auditable and prompt-responsive under the tested contract, but it is not a replay law, repair primitive, or deployable monitor.

Recent policy-family bundle follow-ups sharpen the limit rather than the claim: the current Gemma bundle is fully engaged but not cleanly sufficient, and the current Phi homologous bundle is both EM-only and low-engagement, remaining flat even on the fired rows.

Replay should be kept separate from prompt-gated expression. Clean replay in this paper remains source-side on matched Gemma checkpoints; on Phi, the fixed target is useful for bounded auditing and perturbation tests, not for a stronger control theorem.

Table 12 separates discovery, structural validation, and bounded perturbation testing under the fixed contract used in this paper.

\Needspace{20\baselineskip}
\begingroup
\scriptsize
\setlength{\tabcolsep}{4pt}
\renewcommand{\arraystretch}{1.16}
{\centering
\begin{minipage}{\linewidth}
\captionof{table}{Structural Validation Versus Discovery As Intervention-Target Search}\label{tab:table4-discovery-vs-structural-validation}
\vspace{0.25em}
\centering
\begin{tabular}{>{\RaggedRight\arraybackslash}p{0.12\linewidth}>{\RaggedRight\arraybackslash}p{0.22\linewidth}>{\RaggedRight\arraybackslash}p{0.22\linewidth}>{\RaggedRight\arraybackslash}p{0.20\linewidth}}
\toprule
\textbf{Comparison question} & \textbf{Discovery package} & \textbf{Structural-validation package} & \textbf{Effect on claim} \\
\midrule
What stays fixed? & The same Gemma-\allowbreak{}anchored source-\allowbreak{}defined family, the same target search band, and the same fixed confirmatory set. & The same Gemma-\allowbreak{}anchored source-\allowbreak{}defined family, the same target search band, and the same fixed confirmatory set. & The experiment does not change. \\
Which site is selected? & A broader reasoning-\allowbreak{}side local handle selected for behavioural usefulness and locality. & A narrower late-\allowbreak{}reasoning local lane selected for fidelity to the Gemma-\allowbreak{}anchored source-\allowbreak{}defined family. & Different ranking questions can select different local sites. \\
What is the ranking question? & Which nearby site is most useful as an intervention handle? & Which nearby site is the most faithful target-\allowbreak{}local realisation under structural validation? & Discovery and structural validation answer different scientific questions. \\
What does a positive result support? & A bounded operational handle and an auditable latent target candidate. & A target-\allowbreak{}local realisation with family-\allowbreak{}level correspondence under redistribution. & The same experiment can support one main structural result and one bounded operational consequence. \\
\bottomrule
\end{tabular}
\end{minipage}\par
}
\endgroup

\setcounter{figure}{6}
\begin{figure}[t]
\centering
\includegraphics[width=0.98\linewidth]{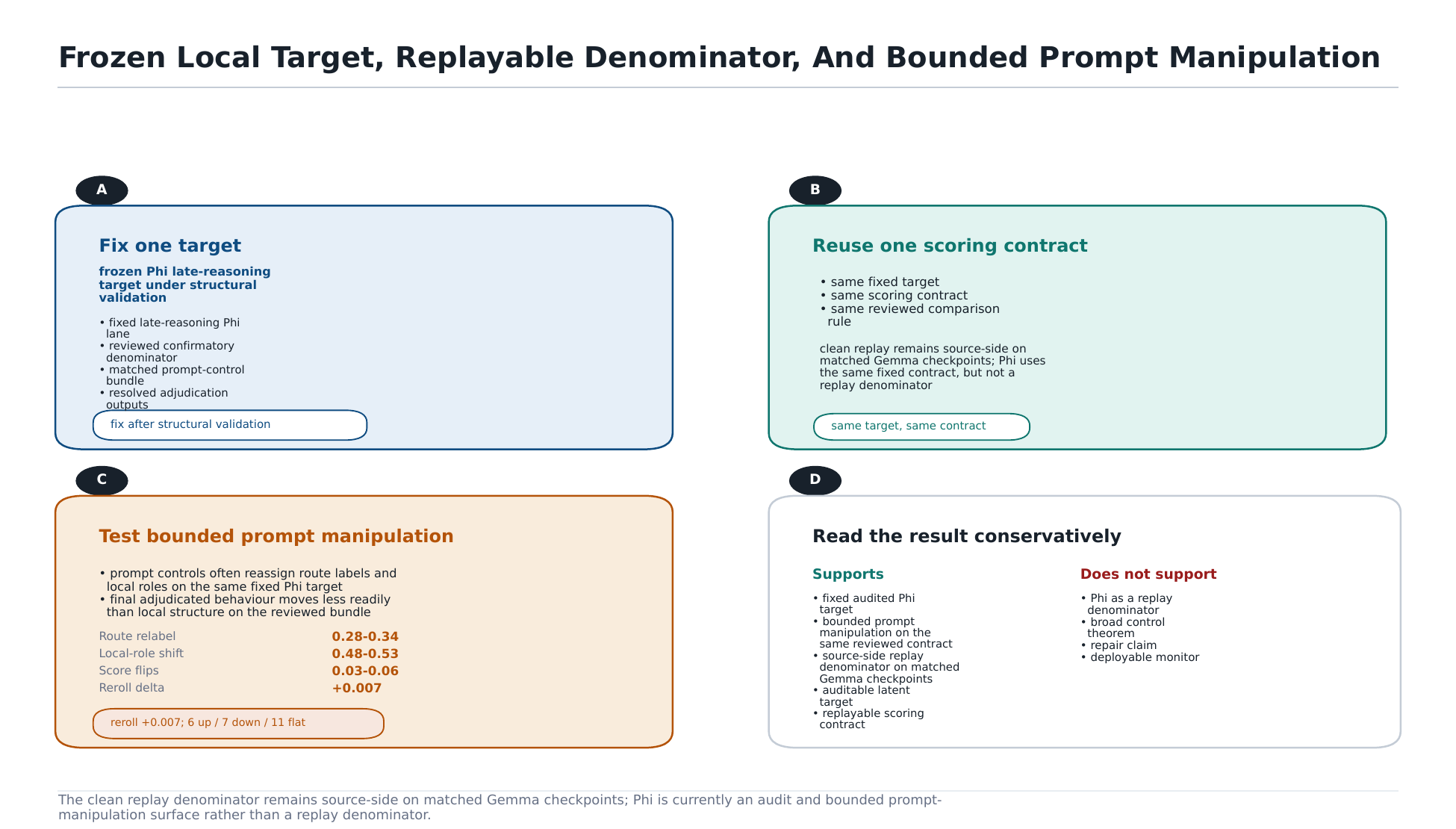}
\caption{Frozen Local Target, Replayable Denominator, And Bounded Prompt Manipulation}
\label{fig:figure7-governance-stack}
\end{figure}
\FloatBarrier

\subsection{Limitations and Future Work}\label{limitations-and-future-work}

The main limitations remain occupancy-faithful closure and loose behaviour coupling. The current evidence does not establish exact cross-system identity, a target-side mediation theorem, compact patch sufficiency, or a deployable monitor. Future work will focus on improvements to our geometry-first constitution-conditioning methods, so that training preserves and actuates the full source-defined family strongly enough for later cross-model and cross-substrate mediation under the same fixed evaluation contract.

\section*{Appendices}\label{appendices}

The appendices separate contract, main retained positives, limiting results, interpretive follow-ups, and provenance. Appendix A states the fixed evaluation contract, Appendix B reports the retained Phi and ALM8 results, Appendix C records the MCQ boundary, Appendix D gathers interpretive follow-ups, and Appendix E indexes provenance and condition translation.

\clearpage
\section*{Appendix A. Experimental Summary, Measurement, And Inclusion Rules}\label{appendix-a.-experimental-summary-measurement-and-inclusion-rules}

Appendix A states which evaluation sets, review rules, and inclusion criteria feed the paper's main claims.

\subsection*{A.0 Experimental Summary And Evidence Map}\label{a.0-experimental-summary-and-evidence-map}

Table 13 provides a compact map of the experimental arc and evidence hierarchy before the appendix detail.

\Needspace{10\baselineskip}
\begin{landscape}
\begingroup
\tiny
\setlength{\tabcolsep}{1pt}
\renewcommand{\arraystretch}{1.12}
\begin{longtable}{>{\RaggedRight\arraybackslash}p{0.07\linewidth}>{\RaggedRight\arraybackslash}p{0.09\linewidth}>{\RaggedRight\arraybackslash}p{0.10\linewidth}>{\RaggedRight\arraybackslash}p{0.10\linewidth}>{\RaggedRight\arraybackslash}p{0.09\linewidth}>{\RaggedRight\arraybackslash}p{0.13\linewidth}>{\RaggedRight\arraybackslash}p{0.12\linewidth}>{\RaggedRight\arraybackslash}p{0.08\linewidth}>{\RaggedRight\arraybackslash}p{0.10\linewidth}}
\caption{Experimental Summary And Evidence Map}\label{tab:table3-evidence-to-claim-matrix}\\
\toprule
\textbf{Stage} & \textbf{System or dataset} & \textbf{What changes} & \textbf{What stays fixed} & \textbf{Main readout} & \textbf{Denominator or protocol} & \textbf{Strongest positive result} & \textbf{Active ceiling} & \textbf{Role in claim ladder} \\
\midrule
\endfirsthead
\toprule
\textbf{Stage} & \textbf{System or dataset} & \textbf{What changes} & \textbf{What stays fixed} & \textbf{Main readout} & \textbf{Denominator or protocol} & \textbf{Strongest positive result} & \textbf{Active ceiling} & \textbf{Role in claim ladder} \\
\midrule
\endhead
\midrule \multicolumn{9}{r}{\emph{Continued on next page}} \\
\endfoot
\bottomrule
\endlastfoot
Gemma Claim 1A & Adapted Gemma 3 1B IT source model & Constitution-\allowbreak{}conditioned post-\allowbreak{}training realises the local source-\allowbreak{}side chart. & Canonical Gemma anchor and the matched adapter-\allowbreak{}ON versus adapter-\allowbreak{}OFF hidden-\allowbreak{}state contrast. & Source-local chart. & Local source anchor plus source-\allowbreak{}side behaviour linkage on the locked source package. & Recoverable source-\allowbreak{}local chart on the canonical Gemma anchor with strong realised-\allowbreak{}row and realised-\allowbreak{}flip support on the locked source package. & This row does not itself close compact exact-\allowbreak{}patch sufficiency. & Claim 1A. \\
Gemma Claim 1B & Adapted Gemma 3 1B IT source model & Broader family-\allowbreak{}support analyses compare nearby local expressions and route completeness across the same source package. & The same locked source package, local anchor, and downstream freeze rule. & Source-\allowbreak{}defined family exported downstream. & Broader family-\allowbreak{}support analyses plus the compact exact-\allowbreak{}patch boundary on the locked source package. & The source-\allowbreak{}local chart sits inside a broader behaviour-\allowbreak{}bearing, route-\allowbreak{}complete family that is the correct downstream contract. & Compact exact-\allowbreak{}patch sufficiency remains unresolved; the portable carrier slice is not the whole source-\allowbreak{}defined family. & Claim 1B. \\
Phi structural-\allowbreak{}validation route & Unadapted Phi-\allowbreak{}4 Mini base model & Only the prompt condition varies; model weights stay fixed. & Frozen Gemma-\allowbreak{}anchored source-\allowbreak{}defined family, predefined six-\allowbreak{}face local band, and the fixed adjudicated confirmatory set. & Target-\allowbreak{}local realisation on a narrow late-\allowbreak{}reasoning lane. & Adjudicated confirmatory free-\allowbreak{}text battery with 128 /\allowbreak{} 128 scored rows and 2688 /\allowbreak{} 2688 control rows reviewed. & Primary reviewed policy result: AUC 0.984 and mean gap 5.502. & Family-\allowbreak{}level correspondence with redistribution; occupancy closure remains incomplete. & Main cross-\allowbreak{}model part of Claim 2. \\
Phi discovery and intervention route & Same Phi-\allowbreak{}4 Mini base model & The ranking protocol shifts from structural fidelity to behavioural usefulness and locality. & Same Gemma-\allowbreak{}anchored source-\allowbreak{}defined family, same fixed confirmatory set, and the same matched, swapped, and no-\allowbreak{}prompt controls. & Intervention handle and auditable latent target. & Search-\allowbreak{}freeze-\allowbreak{}replay workflow on the reviewed Phi prompt-\allowbreak{}control bundle. & Discovery localises a broader reasoning-\allowbreak{}side handle with AUC 0.778 and gap 4.943 that can be frozen and replayed. & Bounded operational consequence only; no broader control or repair claim. & Section 7 bounded operational consequence. \\
MCQ limiting route & Phi multiple-\allowbreak{}choice follow-\allowbreak{}up & Answer-\allowbreak{}inclusive search, subset slicing, hidden-\allowbreak{}state readout, and adjacent no-\allowbreak{}prompt checks are added. & Source-\allowbreak{}family test, reviewed paired comparison, and fixed control comparisons. & Limiting result on reviewed closure. & Reviewed paired free-\allowbreak{}text comparison plus hidden-\allowbreak{}state re-\allowbreak{}entry diagnostic. & A nearby answer-\allowbreak{}side local signal exists, and hidden-\allowbreak{}state re-\allowbreak{}entry is one-\allowbreak{}sided at 110 /\allowbreak{} 192 versus 0 /\allowbreak{} 192. & The stronger reviewed correspondence test does not close; displacement is the dominant failure mode. & Main negative boundary for Claim 2. \\
ALM8 held-\allowbreak{}out corroboration & ALM8 mouse frontal-\allowbreak{}cortex dataset & The frozen Gemma-\allowbreak{}anchored source-\allowbreak{}defined family is tested across held-\allowbreak{}out mouse folds under the ALM8 bridge contract. & Frozen Gemma-\allowbreak{}anchored source-\allowbreak{}defined family, fixed train atlas, exact assigned-\allowbreak{}row matching, and held-\allowbreak{}out scoring. & Held-\allowbreak{}out corroboration plus the structurally cleanest local analogue. & Five held-\allowbreak{}out mouse folds under frozen-\allowbreak{}train-\allowbreak{}atlas assignment with train-\allowbreak{}projector causal scoring. & Mean held-\allowbreak{}out AUC 0.717, mean fold gap 4.503, and positive results across all five folds. & Family-\allowbreak{}level correspondence with redistribution rather than exact replay of one local subspace. & Main biological corroboration in Claim 2. \\
Supporting follow-ups & ALM7, Mouse95, Pyramid, Llama 3.1, Qwen, and nearby native Phi & Model family, substrate, or local-\allowbreak{}role context varies across follow-\allowbreak{}up analyses. & The main interpretation stays subordinate to the retained Phi and ALM8 routes, and the source-\allowbreak{}defined family remains the reference contract. & Role continuity, redistribution, and supporting recurrence. & Supporting or interpretive analyses only; not co-\allowbreak{}equal confirmatory routes. & Family regularity survives additional model and substrate variation, sharpening the family-\allowbreak{}level interpretation. & Supporting or interpretive evidence only; it does not outrank the main Phi or ALM8 results. & Appendix D support and interpretation. \\
\end{longtable}
\endgroup
\end{landscape}

Here, the fully adjudicated confirmatory set means the two fixed Phi confirmatory free-text comparisons taken together: 128 / 128 scored rows and 2688 / 2688 control rows receive complete manual semantic adjudication across the primary and secondary routes. The held-out ALM8 result means the main five-fold structural lane scored under a frozen-train-atlas assignment; when 15 / 15 appears elsewhere, it denotes pooled fold-span summaries across the three main spans. The hidden-state contrast is the adapter-on minus adapter-off hidden-state difference on matched captured completion tokens, used to define the frozen source-defined family before downstream testing.

\subsection*{A.1 Main Inference Units And Confirmatory Sets}\label{a.1-main-inference-units-and-confirmatory-sets}

The sections below state the contract that determines which analyses enter the main inference and which remain supporting.

\begin{longtable}[]{@{}
  >{\RaggedRight\arraybackslash}p{(\linewidth - 8\tabcolsep) * \real{0.2000}}
  >{\RaggedRight\arraybackslash}p{(\linewidth - 8\tabcolsep) * \real{0.2000}}
  >{\RaggedRight\arraybackslash}p{(\linewidth - 8\tabcolsep) * \real{0.2000}}
  >{\RaggedRight\arraybackslash}p{(\linewidth - 8\tabcolsep) * \real{0.2000}}
  >{\RaggedRight\arraybackslash}p{(\linewidth - 8\tabcolsep) * \real{0.2000}}@{}}
\caption{Main Inference Units And Confirmatory Sets}\tabularnewline
\toprule\noalign{}
\begin{minipage}[b]{\linewidth}\RaggedRight
Route
\end{minipage} & \begin{minipage}[b]{\linewidth}\RaggedRight
Unit of inference
\end{minipage} & \begin{minipage}[b]{\linewidth}\RaggedRight
Confirmatory set
\end{minipage} & \begin{minipage}[b]{\linewidth}\RaggedRight
Review or protocol requirement
\end{minipage} & \begin{minipage}[b]{\linewidth}\RaggedRight
Authoritative read
\end{minipage} \\
\midrule\noalign{}
\endfirsthead
\toprule\noalign{}
\begin{minipage}[b]{\linewidth}\RaggedRight
Route
\end{minipage} & \begin{minipage}[b]{\linewidth}\RaggedRight
Unit of inference
\end{minipage} & \begin{minipage}[b]{\linewidth}\RaggedRight
Confirmatory set
\end{minipage} & \begin{minipage}[b]{\linewidth}\RaggedRight
Review or protocol requirement
\end{minipage} & \begin{minipage}[b]{\linewidth}\RaggedRight
Authoritative read
\end{minipage} \\
\midrule\noalign{}
\endhead
\bottomrule\noalign{}
\endlastfoot
Phi structural-validation policy evaluation & response row & 64 total rows: 32 informative-constitution rows and 32 null-random-control rows & manual semantic adjudication required; unresolved rows are excluded from the main structural inference & resolved score sheet plus reviewed summary tables \\
Phi secondary honesty/deception evaluation & response row & 64 total rows: 32 informative-constitution rows and 32 null-random-control rows & same fail-closed manual semantic adjudication rule as the primary policy evaluation & resolved score sheet plus reviewed summary tables \\
Phi paired comparison between the exploratory automated screening route and the manually adjudicated structural-validation analysis & paired confirmatory response row & 128 paired rows with identical prompts, generations, and final labels across both analyses & no new scoring layer; inherits the adjudicated labels from the underlying confirmatory analyses & paired comparison bundle over the already adjudicated analyses \\
ALM8 held-out corroboration & held-out ALM8 fold & 5 held-out folds per analysis span and 15 aggregate fold means across the three main spans & no free-text review; the protocol is fixed by frozen-atlas assignment and exact assigned-row matching & ALM8 held-out bridge summary and per-fold local-subspace rows \\
\end{longtable}

These confirmatory sets are part of the result, not bookkeeping. A strong number on an unconstrained exploratory comparison is not treated as interchangeable with a manually adjudicated number on a fixed confirmatory set.

\subsection*{A.2 Manual Semantic Adjudication And Evidence Priority}\label{a.2-manual-semantic-adjudication-and-evidence-priority}

The main free-text comparison is the fully adjudicated Phi structural-validation analysis, not the exploratory automated screening route. That distinction matters because only the former serves as the confirmatory evaluation.

\begin{longtable}[]{@{}
  >{\RaggedRight\arraybackslash}p{(\linewidth - 6\tabcolsep) * \real{0.2500}}
  >{\RaggedRight\arraybackslash}p{(\linewidth - 6\tabcolsep) * \real{0.2500}}
  >{\RaggedRight\arraybackslash}p{(\linewidth - 6\tabcolsep) * \real{0.2500}}
  >{\RaggedRight\arraybackslash}p{(\linewidth - 6\tabcolsep) * \real{0.2500}}@{}}
\caption{Manual Semantic Adjudication Coverage And Evidence Priority}\tabularnewline
\toprule\noalign{}
\begin{minipage}[b]{\linewidth}\RaggedRight
Analysis
\end{minipage} & \begin{minipage}[b]{\linewidth}\RaggedRight
Review status
\end{minipage} & \begin{minipage}[b]{\linewidth}\RaggedRight
Interpretive weight
\end{minipage} & \begin{minipage}[b]{\linewidth}\RaggedRight
Role in the study
\end{minipage} \\
\midrule\noalign{}
\endfirsthead
\toprule\noalign{}
\begin{minipage}[b]{\linewidth}\RaggedRight
Analysis
\end{minipage} & \begin{minipage}[b]{\linewidth}\RaggedRight
Review status
\end{minipage} & \begin{minipage}[b]{\linewidth}\RaggedRight
Interpretive weight
\end{minipage} & \begin{minipage}[b]{\linewidth}\RaggedRight
Role in the study
\end{minipage} \\
\midrule\noalign{}
\endhead
\bottomrule\noalign{}
\endlastfoot
Fully adjudicated Phi structural-validation analysis & fully adjudicated: 128 / 128 scored rows and 2688 / 2688 control rows receive complete manual semantic adjudication & main evidence for the Phi result & canonical Phi result \\
Exploratory automated screening route & partial manual semantic adjudication coverage only: 41 / 64 scored rows and 861 / 1344 control rows receive review; the remainder rely on automated prefix rules & screening and comparator only, not co-equal main evidence & discovery and control comparator \\
Paired comparison between exploratory screening and adjudicated analyses & paired rows preserve paired question identifiers, same-question grouping, condition role, prompts, generations, and final labels & valid paired comparison & shows that the shift in selected local subspace is not a row-identity artefact \\
ALM8 held-out bridge & fixed held-out fold protocol with zero missing expected rows after compliance reconciliation & main biological corroboration at the same conservative interpretation & cross-substrate corroboration at the same conservative interpretation \\
\end{longtable}

Manual semantic adjudication also materially changes the scientific picture. On the Phi confirmatory routes, manual semantic adjudication changed 41 primary-policy labels and 32 secondary-evaluation labels relative to the exploratory automated screening surface. The study therefore treats the adjudicated semantic labels, not the automated screening outputs, as authoritative wherever free-text behaviour enters the main structural inference.

On the paired comparison bundle, 0 / 128 rows change binary score or final label; what changes is the continuous readout surface, whose paired signed-score delta has mean -0.55, bootstrap interval {[}-0.99, -0.12{]}, and sign-flip permutation p = 0.013.

The released adjudication pack also shows why this matters: one case is a genuine deception-to-refusal semantic reversal, while another is a strict prefix inversion in which the safer answer receives the lower automatic score. Manual semantic adjudication is therefore evidence-defining, not a cosmetic scorer correction.

\subsection*{A.3 Condition Labels}\label{a.3-condition-labels}

The canonical family contrast is reported constitution-first rather than by arm labels alone. That rule prevents the structural conclusion from drifting into an unsupported generic low-versus-high wording. Throughout the appendix suite, the shorter labels below are used after the exact condition IDs are given once.

\begin{longtable}[]{@{}
  >{\RaggedRight\arraybackslash}p{(\linewidth - 4\tabcolsep) * \real{0.3333}}
  >{\RaggedRight\arraybackslash}p{(\linewidth - 4\tabcolsep) * \real{0.3333}}
  >{\RaggedRight\arraybackslash}p{(\linewidth - 4\tabcolsep) * \real{0.3333}}@{}}
\caption{Resolved Condition Labels Used In Prose}\tabularnewline
\toprule\noalign{}
\begin{minipage}[b]{\linewidth}\RaggedRight
Operational condition id
\end{minipage} & \begin{minipage}[b]{\linewidth}\RaggedRight
Resolved condition identity
\end{minipage} & \begin{minipage}[b]{\linewidth}\RaggedRight
Short label used in prose
\end{minipage} \\
\midrule\noalign{}
\endfirsthead
\toprule\noalign{}
\begin{minipage}[b]{\linewidth}\RaggedRight
Operational condition id
\end{minipage} & \begin{minipage}[b]{\linewidth}\RaggedRight
Resolved condition identity
\end{minipage} & \begin{minipage}[b]{\linewidth}\RaggedRight
Short label used in prose
\end{minipage} \\
\midrule\noalign{}
\endhead
\bottomrule\noalign{}
\endlastfoot
\texttt{low\_\allowbreak{}si\_\allowbreak{}on} & \texttt{constitutions-\allowbreak{}v2-\allowbreak{}high\_\allowbreak{}effective\_\allowbreak{}mi} & informative constitution \\
\texttt{high\_\allowbreak{}si\_\allowbreak{}on} & \texttt{null\_\allowbreak{}random\_\allowbreak{}v1} & null-random control \\
\end{longtable}

The operational arm names are historical and are not used as scientific labels in the study. Throughout Appendix A-D, the canonical family contrast is therefore the informative constitution (\texttt{constitutions-\allowbreak{}v2-\allowbreak{}high\_\allowbreak{}effective\_\allowbreak{}mi}) versus the null-random control (\texttt{null\_\allowbreak{}random\_\allowbreak{}v1}) under source-conditioned retargeting. The point is not to generalise to a universal low-versus-high selectivity axis. Phase migration, output-linked expression, and nearby native correspondence may depend on this resolved condition pair rather than on the arm labels themselves.

\subsection*{A.4 Evidence Hierarchy And Scientific Weight}\label{a.4-evidence-hierarchy-and-scientific-weight}

The evidence base spans source-chart evidence, main Phi evidence, main biological corroboration on ALM8, limiting results, and supporting follow-ups. The table below states how those lines are used in the paper.

\begin{longtable}[]{@{}
  >{\RaggedRight\arraybackslash}p{(\linewidth - 6\tabcolsep) * \real{0.2500}}
  >{\RaggedRight\arraybackslash}p{(\linewidth - 6\tabcolsep) * \real{0.2500}}
  >{\RaggedRight\arraybackslash}p{(\linewidth - 6\tabcolsep) * \real{0.2500}}
  >{\RaggedRight\arraybackslash}p{(\linewidth - 6\tabcolsep) * \real{0.2500}}@{}}
\caption{Evidence Hierarchy And Scientific Weight}\tabularnewline
\toprule\noalign{}
\begin{minipage}[b]{\linewidth}\RaggedRight
Evidence level
\end{minipage} & \begin{minipage}[b]{\linewidth}\RaggedRight
Included analyses
\end{minipage} & \begin{minipage}[b]{\linewidth}\RaggedRight
Manuscript use
\end{minipage} & \begin{minipage}[b]{\linewidth}\RaggedRight
Scientific weight
\end{minipage} \\
\midrule\noalign{}
\endfirsthead
\toprule\noalign{}
\begin{minipage}[b]{\linewidth}\RaggedRight
Evidence level
\end{minipage} & \begin{minipage}[b]{\linewidth}\RaggedRight
Included analyses
\end{minipage} & \begin{minipage}[b]{\linewidth}\RaggedRight
Manuscript use
\end{minipage} & \begin{minipage}[b]{\linewidth}\RaggedRight
Scientific weight
\end{minipage} \\
\midrule\noalign{}
\endhead
\bottomrule\noalign{}
\endlastfoot
Source-chart evidence & source-side Gemma local-chart analyses, broader family-support analyses, compact exact-patch boundary, and source-face provenance audit & main text Sections 1 and 3 & directly supports Claim 1A and Claim 1B on Gemma while limiting stronger source-side claims \\
Main cross-model evidence & reviewed Phi layer-24 reasoning target-local realisation on the fixed fully adjudicated confirmatory set, together with paired prompt-control comparisons and complete reviewed semantic adjudication & main-text quantitative tables and direct scientific language & directly supports the main Phi result \\
Main biological corroboration & ALM8 held-out bridge under the fixed frozen-atlas protocol & main-text biological corroboration at the same conservative interpretation & supports the same cross-substrate interpretation without turning it into exact identity \\
Limiting evidence & multiple-choice limiting analysis and occupancy-focused closure diagnostics & main-text limits language and Appendix C analysis & sets the limit of inference on stronger source-to-target correspondence claims \\
Supporting biological lines & Mouse95 and ALM7 follow-ups \cite{li2022alm7dataset}, together with Pyramid \cite{musall2023pyramidal,churchland2023pyramiddata} & Appendix D and discussion support & clarify role continuity, redistribution, and deformation without outranking ALM8 \\
Supporting cross-model lines & Llama 3.1 8B Instruct \cite{meta2024llama31_8b_instruct}, Qwen3 8B \cite{qwen2025qwen3_8b}, and nearby target-native Phi-4 Mini Instruct corroboration \cite{microsoft2025phi4miniinstruct} & Appendix D and discussion support & test whether the broader family-level pattern extends beyond the main Phi result without widening the main supported interpretation \\
Interpretive follow-up analyses & local-family audit, route-level replay, transition-level replay, leave-one-out ablation, occupancy-first recalculation, and broader role-family synthesis & Appendix D and discussion support & refines ontology and failure interpretation, but does not outrank the main Phi or ALM8 results \\
\end{longtable}

This hierarchy is essential to the study's logic. Within the source-side chart level, the paper keeps three source-side components separate whenever claim language matters: the local source anchor, the broader source-side family-support analyses, and the compact exact-patch boundary. The broader interpretive follow-ups support a shift from single-subspace language toward a narrow local family with differentiated member roles, but they do not elevate a broader route-level or hidden-family intervention above the reviewed Phi result or ALM8 held-out corroboration. With that inclusion rule fixed, Appendix B records the main Phi result and the main ALM8 corroboration result.

\clearpage
\section*{Appendix B. Main Phi Structural Result And Main ALM8 Held-Out Corroboration}\label{appendix-b.-main-phi-structural-result-and-main-alm8-held-out-corroboration}

Appendix B reports the main retained Phi and ALM8 quantitative results under the Appendix A contract.

\subsection*{B.1 Reviewed Phi Structural-Validation Summary}\label{b.1-reviewed-phi-structural-validation-summary}

This section collects the main retained Phi numbers before the structural-interpretation detail and ALM8 bridge tables.

\begin{longtable}[]{@{}
  >{\RaggedRight\arraybackslash}p{(\linewidth - 4\tabcolsep) * \real{0.3333}}
  >{\RaggedRight\arraybackslash}p{(\linewidth - 4\tabcolsep) * \real{0.3333}}
  >{\RaggedRight\arraybackslash}p{(\linewidth - 4\tabcolsep) * \real{0.3333}}@{}}
\caption{Reviewed Phi Structural-Validation Summary}\tabularnewline
\toprule\noalign{}
\begin{minipage}[b]{\linewidth}\RaggedRight
Quantity
\end{minipage} & \begin{minipage}[b]{\linewidth}\RaggedRight
Value
\end{minipage} & \begin{minipage}[b]{\linewidth}\RaggedRight
Why it matters
\end{minipage} \\
\midrule\noalign{}
\endfirsthead
\toprule\noalign{}
\begin{minipage}[b]{\linewidth}\RaggedRight
Quantity
\end{minipage} & \begin{minipage}[b]{\linewidth}\RaggedRight
Value
\end{minipage} & \begin{minipage}[b]{\linewidth}\RaggedRight
Why it matters
\end{minipage} \\
\midrule\noalign{}
\endhead
\bottomrule\noalign{}
\endlastfoot
Selected target-local realisation & \texttt{reason} face in Phi layer 24 on the hidden-state contrast surface & corrected canonical structural-validation subspace on Phi \\
Internal search index & 5 & reproducibility note for the frozen local face \\
Frozen-face summary score & 123.82 & witness-first retained face on the corrected rerun \\
Discovery axis norm & 3.96 & nontrivial local separation at the selected local subspace \\
Patch coverage fraction mean & not materialised & the corrected retained face has no attached patch-coverage summary \\
Primary policy route reviewed denominator and uncertainty & \texttt{64} reviewed rows from the fixed primary route; AUC \texttt{0.984} (bootstrap \texttt{95\%} CI \texttt{{[}0.952,\ 1.000{]}}); mean gap \texttt{5.50} (\texttt{95\%} CI \texttt{{[}4.87,\ 6.17{]}}); label-permutation \texttt{p\ =\allowbreak{}\ 0.03125} & strongest reviewed behavioural separation in the main Phi result \\
Primary route control placement & above the maxima reached by the \texttt{21} materialised confirmatory null, random, and orthogonal control surfaces (AUC max \texttt{0.669}, mean-gap max \texttt{1.17}), equivalent to the \texttt{100th} percentile of both control ranges & the headline Phi number is control-clean rather than an isolated point estimate \\
Informative-constitution mean signed score & +1.47 & positive branch on the informative constitution \\
Null-random-control mean signed score & -2.06 & negative branch on the null-random control \\
Secondary evaluation reviewed denominator and uncertainty & \texttt{64} reviewed rows from the fixed secondary route; AUC \texttt{0.622} (bootstrap \texttt{95\%} CI \texttt{{[}0.441,\ 0.785{]}}); mean gap \texttt{1.13} (\texttt{95\%} CI \texttt{{[}-1.26,\ 3.29{]}}); label-permutation \texttt{p\_\allowbreak{}auc\ =\allowbreak{}\ 0.0966}, \texttt{p\_\allowbreak{}gap\ =\allowbreak{}\ 0.1873} & weaker but still positive support on the same frozen face \\
Secondary evaluation aligned counts and control status & \texttt{30\ /\allowbreak{}\ 32} informative-constitution rows versus \texttt{22\ /\allowbreak{}\ 32} null-random-control rows aligned; does not clear the strongest same-rank confirmatory control surfaces & why the secondary route remains supporting rather than co-equal evidence \\
\end{longtable}

These denominator-first summaries are the quantitative core of the main Phi result. They establish that the selected target-local Phi subspace is real, behaviour-linked, and strongly branch-separating on the reviewed policy evaluation while remaining transparent about uncertainty and control placement.

\subsection*{B.2 Structural Interpretation And Nearby Target-Native Corroboration}\label{b.2-structural-interpretation-and-nearby-target-native-corroboration}

\begin{longtable}[]{@{}
  >{\RaggedRight\arraybackslash}p{(\linewidth - 4\tabcolsep) * \real{0.3333}}
  >{\RaggedRight\arraybackslash}p{(\linewidth - 4\tabcolsep) * \real{0.3333}}
  >{\RaggedRight\arraybackslash}p{(\linewidth - 4\tabcolsep) * \real{0.3333}}@{}}
\caption{Phi Structural Interpretation And Nearby Target-Native Corroboration}\tabularnewline
\toprule\noalign{}
\begin{minipage}[b]{\linewidth}\RaggedRight
Structural quantity
\end{minipage} & \begin{minipage}[b]{\linewidth}\RaggedRight
Value
\end{minipage} & \begin{minipage}[b]{\linewidth}\RaggedRight
Interpretation
\end{minipage} \\
\midrule\noalign{}
\endfirsthead
\toprule\noalign{}
\begin{minipage}[b]{\linewidth}\RaggedRight
Structural quantity
\end{minipage} & \begin{minipage}[b]{\linewidth}\RaggedRight
Value
\end{minipage} & \begin{minipage}[b]{\linewidth}\RaggedRight
Interpretation
\end{minipage} \\
\midrule\noalign{}
\endhead
\bottomrule\noalign{}
\endlastfoot
Structural interpretation used in the study & family-level structural correspondence with redistribution & positive correspondence below exact identity \\
Exact identity support & false & exact-identity language is not supported \\
Chart-level support outcome & pass & local structural support is present \\
Occupancy support outcome & blocked & occupancy-faithful closure fails \\
Structural support score & 0.56 & moderate structural support on the canonical policy evaluation \\
Main limiting term & occupancy-distance term (\texttt{occ\_\allowbreak{}w2\_\allowbreak{}sq\_\allowbreak{}norm}) & occupancy redistribution limits the stronger identity claim \\
Nearby target-native regime & late-reasoning hidden-state readout around Phi layer 28 & independent nearby target-native corroboration \\
Relation to the selected retargeted subspace & adjacent phase regime & target-local realisation rather than literal source-basis copying \\
Base-hidden score & 5437.46 & the nearby native regime is strong and nontrivial \\
Base-hidden captured fraction & 0.32 & target-native corroboration is partial and localised \\
Base-hidden row summary & 6 success rows, 1 control row & nearby native activity exists without collapsing the retargeted and native sites into one exact local subspace \\
\end{longtable}

This is why the preferred interpretation is family-level structural correspondence with redistribution. The Phi result is stronger than a null or noise account, but it remains below exact occupancy-faithful identity.

\subsection*{B.3 ALM8 Held-Out Corroboration Under The Same Conservative Interpretation}\label{b.3-alm8-held-out-corroboration-under-the-same-conservative-interpretation}

\begin{longtable}[]{@{}
  >{\RaggedRight\arraybackslash}p{(\linewidth - 4\tabcolsep) * \real{0.3333}}
  >{\RaggedRight\arraybackslash}p{(\linewidth - 4\tabcolsep) * \real{0.3333}}
  >{\RaggedRight\arraybackslash}p{(\linewidth - 4\tabcolsep) * \real{0.3333}}@{}}
\caption{ALM8 Held-Out Corroboration Under Conservative Interpretation}\tabularnewline
\toprule\noalign{}
\begin{minipage}[b]{\linewidth}\RaggedRight
Quantity
\end{minipage} & \begin{minipage}[b]{\linewidth}\RaggedRight
Value
\end{minipage} & \begin{minipage}[b]{\linewidth}\RaggedRight
Interpretation
\end{minipage} \\
\midrule\noalign{}
\endfirsthead
\toprule\noalign{}
\begin{minipage}[b]{\linewidth}\RaggedRight
Quantity
\end{minipage} & \begin{minipage}[b]{\linewidth}\RaggedRight
Value
\end{minipage} & \begin{minipage}[b]{\linewidth}\RaggedRight
Interpretation
\end{minipage} \\
\midrule\noalign{}
\endhead
\bottomrule\noalign{}
\endlastfoot
Unit of inference & held-out ALM8 fold & fixed cross-substrate protocol \\
Structurally cleanest structural analogue & late-reasoning readout-adjacent bridge face & structurally cleanest held-out bridge subspace \\
Mean held-out AUC & 0.72 & positive cross-substrate separation \\
Mean fold gap & 4.50 & substantial held-out effect size \\
Positive fold fraction & 1.00 & every held-out fold remains positive \\
Held-out structural-validation support fraction & 0.80 & strict structural support is present on most folds \\
Held-out chart-level support fraction & 1.00 & chart-level support is stable \\
Held-out occupancy support fraction & 0.80 & occupancy support is stronger here than on the model-side Phi result \\
Held-out structural support score mean & 0.63 & corroborative structural strength \\
Structural interpretation across held-out folds & family-level structural correspondence with redistribution in all 5 / 5 held-out folds & supports family-level recurrence rather than exact identity \\
\end{longtable}

The ALM8 result therefore supports the same level of structural inference as the Phi policy result. It strengthens the recurrence result by surviving cross-substrate variation, but it does not upgrade the inference into exact local-subspace identity.

This bridge should be read at the level of the held-out bridge rather than as a long list of independent recurrences, because the current ALM8 evidence is pooled across closely related bridge instances.

On the structurally cleanest late-reasoning lane, the held-out fold gap carries a 95\% bootstrap interval {[}+1.99, +8.22{]} and an exact one-sided sign test of 5 / 5 positive (p = 0.031), so the positive read is not carried by one anomalous fold.

\subsection*{B.4 Why ALM8 Requires Two Readouts}\label{b.4-why-alm8-requires-two-readouts}

The ALM8 bridge answers two different ranking questions. That split is scientifically useful because it distinguishes raw predictive strength from the cleaner structural analogue.

\begin{longtable}[]{@{}
  >{\RaggedRight\arraybackslash}p{(\linewidth - 6\tabcolsep) * \real{0.2500}}
  >{\RaggedRight\arraybackslash}p{(\linewidth - 6\tabcolsep) * \real{0.2500}}
  >{\RaggedRight\arraybackslash}p{(\linewidth - 6\tabcolsep) * \real{0.2500}}
  >{\RaggedRight\arraybackslash}p{(\linewidth - 6\tabcolsep) * \real{0.2500}}@{}}
\caption{Why ALM8 Requires Two Readouts}\tabularnewline
\toprule\noalign{}
\begin{minipage}[b]{\linewidth}\RaggedRight
Comparison question
\end{minipage} & \begin{minipage}[b]{\linewidth}\RaggedRight
Preferred span
\end{minipage} & \begin{minipage}[b]{\linewidth}\RaggedRight
Key metrics
\end{minipage} & \begin{minipage}[b]{\linewidth}\RaggedRight
What it means
\end{minipage} \\
\midrule\noalign{}
\endfirsthead
\toprule\noalign{}
\begin{minipage}[b]{\linewidth}\RaggedRight
Comparison question
\end{minipage} & \begin{minipage}[b]{\linewidth}\RaggedRight
Preferred span
\end{minipage} & \begin{minipage}[b]{\linewidth}\RaggedRight
Key metrics
\end{minipage} & \begin{minipage}[b]{\linewidth}\RaggedRight
What it means
\end{minipage} \\
\midrule\noalign{}
\endhead
\bottomrule\noalign{}
\endlastfoot
Which span does the native held-out system prefer in its own discovery analysis? & answer & native train peak counts: 15 / 15 answer, 0 / 15 late-reasoning, 0 / 15 reasoning & native biological expression is answer-heavy \\
Which span has the strongest raw held-out bridge score? & reasoning & mean fold gap +6.39, mean AUC 0.85, positive folds 15 / 15 & raw bridge strength favours a broader output-linked expression \\
Which span best satisfies the structural-validation protocol? & late reasoning & mean fold gap +4.50, mean AUC 0.72, structural-validation support 0.80, structural support score 0.63 & structural corroboration favours the structurally cleanest structural analogue near the readout \\
\end{longtable}

This is the clearest interpretation of ALM8. Cross-substrate corroboration is real, but local expression shifts outward toward more output-linked observables. The positive result is therefore recurrence with redistribution, not exact replay of one fixed local subspace. Appendix C then records the strongest limits on that result.

\clearpage
\section*{Appendix C. Limiting Results And Closure Diagnostics}\label{appendix-c.-limiting-results-and-closure-diagnostics}

Appendix C records the strongest negative boundary in the paper: MCQ localises a nearby signal without closing the stronger correspondence test.

\subsection*{C.1 Multiple-Choice Limiting Analysis}\label{c.1-multiple-choice-limiting-analysis}

The multiple-choice analysis is the strongest limiting result in the appendix suite. The table below shows the shortest route to the boundary: the main rescue routes were executed, and none recovered a behaviourally specific structural-validation result on the reviewed paired denominator.

\begin{longtable}[]{@{}
  >{\RaggedRight\arraybackslash}p{(\linewidth - 6\tabcolsep) * \real{0.2500}}
  >{\RaggedRight\arraybackslash}p{(\linewidth - 6\tabcolsep) * \real{0.2500}}
  >{\RaggedRight\arraybackslash}p{(\linewidth - 6\tabcolsep) * \real{0.2500}}
  >{\RaggedRight\arraybackslash}p{(\linewidth - 6\tabcolsep) * \real{0.2500}}@{}}
\caption{MCQ Limiting Analysis: Rescue Attempts And Failure Modes}\tabularnewline
\toprule\noalign{}
\begin{minipage}[b]{\linewidth}\RaggedRight
Correction or diagnostic step
\end{minipage} & \begin{minipage}[b]{\linewidth}\RaggedRight
What it tested
\end{minipage} & \begin{minipage}[b]{\linewidth}\RaggedRight
Main outcome
\end{minipage} & \begin{minipage}[b]{\linewidth}\RaggedRight
Boundary contribution
\end{minipage} \\
\midrule\noalign{}
\endfirsthead
\toprule\noalign{}
\begin{minipage}[b]{\linewidth}\RaggedRight
Correction or diagnostic step
\end{minipage} & \begin{minipage}[b]{\linewidth}\RaggedRight
What it tested
\end{minipage} & \begin{minipage}[b]{\linewidth}\RaggedRight
Main outcome
\end{minipage} & \begin{minipage}[b]{\linewidth}\RaggedRight
Boundary contribution
\end{minipage} \\
\midrule\noalign{}
\endhead
\bottomrule\noalign{}
\endlastfoot
Source-provenance correction & whether the failure was caused by the wrong source lineage & source correction did not rescue the test & failure is not explained by missing source provenance alone \\
Answer-inclusive rerun & whether the original Phi search band missed the relevant target span & the selected local subspace moved to layer 24 on the answer span, but the paired source-to-target comparison remained negative & failure is not explained by a simple search-band omission \\
Fixed intervention comparison & whether a detectable target-local subspace automatically yields a causal localisation result & the selected answer-side subspace remained behaviourally reactive but did not beat the null-random control, random, or orthogonal controls & local behavioural sensitivity is not enough for the stronger correspondence claim \\
Subset slicing & whether a cleaner subset on the paired denominator would reveal a behaviourally specific signal on the primary policy evaluation & no principled subset produced a convincing result on the reviewed paired comparison & there is no supported narrow-subset rescue \\
Hidden-state source-to-target diagnostic & whether a nearby hidden-state local subspace closes the correspondence question even if the delta-space test fails & a nearby hidden-state local subspace exists, but frozen re-entry into the source-family test is one-sided & localisation and exact closure are distinct \\
Adjacent no-prompt control & whether Phi-side constitution text is incidental once the nearby local family has been found & removing only Phi-side constitution text collapses discovery differential and confirmatory arm separation into a shared local routine & prompt-side branch dependence is real even when nearby local structure exists \\
\end{longtable}

\subsection*{C.2 Representative Negative And Near-Positive MCQ Evaluations}\label{c.2-representative-negative-and-near-positive-mcq-evaluations}

Unless stated otherwise, the MCQ rows below refer to the full reviewed matched-family paired denominator of \texttt{512} rows.

\begin{longtable}[]{@{}
  >{\RaggedRight\arraybackslash}p{(\linewidth - 6\tabcolsep) * \real{0.2500}}
  >{\RaggedRight\arraybackslash}p{(\linewidth - 6\tabcolsep) * \real{0.2500}}
  >{\RaggedRight\arraybackslash}p{(\linewidth - 6\tabcolsep) * \real{0.2500}}
  >{\RaggedRight\arraybackslash}p{(\linewidth - 6\tabcolsep) * \real{0.2500}}@{}}
\caption{Representative Negative And Near-Positive MCQ Evaluations}\tabularnewline
\toprule\noalign{}
\begin{minipage}[b]{\linewidth}\RaggedRight
Analysis case
\end{minipage} & \begin{minipage}[b]{\linewidth}\RaggedRight
Behavioural read
\end{minipage} & \begin{minipage}[b]{\linewidth}\RaggedRight
Structural read
\end{minipage} & \begin{minipage}[b]{\linewidth}\RaggedRight
Why it does not support the main structural result
\end{minipage} \\
\midrule\noalign{}
\endfirsthead
\toprule\noalign{}
\begin{minipage}[b]{\linewidth}\RaggedRight
Analysis case
\end{minipage} & \begin{minipage}[b]{\linewidth}\RaggedRight
Behavioural read
\end{minipage} & \begin{minipage}[b]{\linewidth}\RaggedRight
Structural read
\end{minipage} & \begin{minipage}[b]{\linewidth}\RaggedRight
Why it does not support the main structural result
\end{minipage} \\
\midrule\noalign{}
\endhead
\bottomrule\noalign{}
\endlastfoot
Original reasoning-side delta subspace & full paired-condition confirmatory comparison: AUC = 0.43, mean gap = -0.49 & family-level structural correspondence with redistribution; chart-level support passes, occupancy support fails & remains wrong-signed in the reviewed paired comparison \\
Answer-inclusive answer-side delta subspace & full paired-condition confirmatory comparison: AUC = 0.45, mean gap = -0.34 & family-level structural correspondence with redistribution; chart-level support passes, occupancy support fails & still loses to stronger random and orthogonal controls \\
Answer-side high-confidence paired subset & nominally positive behavioural slice: AUC = 0.53, mean gap = 0.26 & broader family-level correspondence; both chart-level and occupancy support fail & turns positive only by giving up the structural interpretation needed for the stronger source-to-target correspondence claim \\
Hidden reasoning-rich comparison subset & superficially positive behavioural slice: AUC = 0.60, mean gap = 5.72 & no paired-condition counterpart is available on the completed one-condition hidden analysis & not on the same reviewed paired comparison \\
\end{longtable}

These rows clarify the shape of the failure. The MCQ analysis is not uniformly null. It contains localised and even behaviourally nontrivial target-side structure. What it does not contain is the full combination of paired behavioural separation, control-clean specificity, and structural closure in the same reviewed evaluation.

\subsection*{C.3 Hidden-State Source-To-Target Diagnostic}\label{c.3-hidden-state-source-to-target-diagnostic}

The hidden-state chain adds an important distinction between target-side localisation and true source-to-target closure. In hidden space, the multiple-choice analysis does recover a nearby target local subspace on the canonical recurrent-support band. What fails is symmetric re-entry into the frozen source-family test.

\begin{longtable}[]{@{}
  >{\RaggedRight\arraybackslash}p{(\linewidth - 4\tabcolsep) * \real{0.3333}}
  >{\RaggedRight\arraybackslash}p{(\linewidth - 4\tabcolsep) * \real{0.3333}}
  >{\RaggedRight\arraybackslash}p{(\linewidth - 4\tabcolsep) * \real{0.3333}}@{}}
\caption{Hidden-State Source-To-Target Diagnostic Summary}\tabularnewline
\toprule\noalign{}
\begin{minipage}[b]{\linewidth}\RaggedRight
Hidden-state diagnostic
\end{minipage} & \begin{minipage}[b]{\linewidth}\RaggedRight
Value
\end{minipage} & \begin{minipage}[b]{\linewidth}\RaggedRight
Interpretation
\end{minipage} \\
\midrule\noalign{}
\endfirsthead
\toprule\noalign{}
\begin{minipage}[b]{\linewidth}\RaggedRight
Hidden-state diagnostic
\end{minipage} & \begin{minipage}[b]{\linewidth}\RaggedRight
Value
\end{minipage} & \begin{minipage}[b]{\linewidth}\RaggedRight
Interpretation
\end{minipage} \\
\midrule\noalign{}
\endhead
\bottomrule\noalign{}
\endlastfoot
Strongest localised target subspace & layer 24, reason, \texttt{hidden\_\allowbreak{}on} & nearby target local subspace exists on the canonical band \\
Nearby supporting subspaces & layer 24 late-reasoning, layer 23 reasoning, layer 23 late-reasoning & the correspondence is local and structured rather than isolated noise \\
Informative-constitution frozen re-entry & 110 / 192 = 0.57 & the informative-constitution branch re-enters the source atlas on a substantial fraction of rows \\
Null-random-control frozen re-entry & 0 / 192 = 0.00 & the null-random-control branch does not re-enter the source atlas at all \\
Strict replay closure & 0 strict family or aggregate replay rows & there is no replay-closure result that would raise the inference to exact source-to-target closure \\
Dominant rejection reason & 268 centroid-distance failures versus 6 basis-angle failures & failure is governed primarily by displacement rather than by simple rotation \\
Mean normalised centroid distance & informative constitution 3.48, null-random control 7.10 & the null-random-control branch lands far outside the source slot scale \\
Mean normalised angle & informative constitution 2.04, null-random control 2.25 & angle mismatch exists, but it is not the dominant failure mode \\
\end{longtable}

This is why the hidden MCQ result is scientifically useful even though it remains below exact source-to-target closure. It shows that a nearby target-hidden local subspace can be present while exact re-entry into a frozen source atlas remains one-sided and constitution-sensitive.

\subsection*{C.4 Occupancy And Closure Diagnostics On The Policy Family}\label{c.4-occupancy-and-closure-diagnostics-on-the-policy-family}

The main Phi result also has a clear limit. An occupancy-focused recalculation on the reviewed family comparison shows that the limit of inference is not mainly a tangent-rank artefact or a weak structural support score. The limiting factor is how differently the informative constitution and null-random control occupy the shared recurrent support.

\begin{longtable}[]{@{}
  >{\RaggedRight\arraybackslash}p{(\linewidth - 4\tabcolsep) * \real{0.3333}}
  >{\RaggedRight\arraybackslash}p{(\linewidth - 4\tabcolsep) * \real{0.3333}}
  >{\RaggedRight\arraybackslash}p{(\linewidth - 4\tabcolsep) * \real{0.3333}}@{}}
\caption{Occupancy And Closure Diagnostics On The Policy Family}\tabularnewline
\toprule\noalign{}
\begin{minipage}[b]{\linewidth}\RaggedRight
Diagnostic comparison
\end{minipage} & \begin{minipage}[b]{\linewidth}\RaggedRight
Main numbers
\end{minipage} & \begin{minipage}[b]{\linewidth}\RaggedRight
What it shows
\end{minipage} \\
\midrule\noalign{}
\endfirsthead
\toprule\noalign{}
\begin{minipage}[b]{\linewidth}\RaggedRight
Diagnostic comparison
\end{minipage} & \begin{minipage}[b]{\linewidth}\RaggedRight
Main numbers
\end{minipage} & \begin{minipage}[b]{\linewidth}\RaggedRight
What it shows
\end{minipage} \\
\midrule\noalign{}
\endhead
\bottomrule\noalign{}
\endlastfoot
Late-reasoning corridor, same-site informative constitution versus null-random control & \texttt{occ\_\allowbreak{}w2\_\allowbreak{}sq\_\allowbreak{}norm} = 1.02-1.07, \texttt{energy\_\allowbreak{}distance\_\allowbreak{}norm} = 0.23-0.24 & the best local corridor still fails occupancy closure by a wide margin \\
Reasoning corridor, same-site informative constitution versus null-random control & \texttt{occ\_\allowbreak{}w2\_\allowbreak{}sq\_\allowbreak{}norm} = 1.72-1.78, \texttt{energy\_\allowbreak{}distance\_\allowbreak{}norm} = 1.43-1.46 & broader reasoning subspaces are markedly less occupancy-faithful than late reasoning \\
Canonical occupancy thresholds & \texttt{max\_\allowbreak{}occ\_\allowbreak{}w2\_\allowbreak{}sq\_\allowbreak{}norm} = 0.55, \texttt{max\_\allowbreak{}energy\_\allowbreak{}distance\_\allowbreak{}norm} = 0.65 & none of the same-site informative-constitution-versus-null-random-control comparisons are close enough to pass \\
Same-span control pair: layer 24 late-reasoning versus layer 23 late-reasoning & \texttt{occ\_\allowbreak{}w2\_\allowbreak{}sq\_\allowbreak{}norm} = 0.0166, \texttt{energy\_\allowbreak{}distance\_\allowbreak{}norm} = 0.0043 & the structural-validation procedure can detect tight local closure when closure is present \\
Same-span control pair: layer 25 reasoning versus layer 24 reasoning & \texttt{occ\_\allowbreak{}w2\_\allowbreak{}sq\_\allowbreak{}norm} = 0.0195, \texttt{energy\_\allowbreak{}distance\_\allowbreak{}norm} = 0.0106 & nearby members within one corridor can be tightly matched \\
Same-span control pair: layer 25 reasoning versus layer 23 reasoning & \texttt{occ\_\allowbreak{}w2\_\allowbreak{}sq\_\allowbreak{}norm} = 0.0761, \texttt{energy\_\allowbreak{}distance\_\allowbreak{}norm} = 0.0286 & even broader same-corridor pairs remain far tighter than the paired informative-constitution-versus-null-random-control comparison \\
\end{longtable}

The resulting interpretation is straightforward. The local family is real and internally coherent. The stronger identity-style closure fails because the informative constitution and null-random control populate that shared recurrent support differently enough that occupancy support does not currently close.

\subsection*{C.5 Limiting Statement}\label{c.5-limiting-statement}

Taken together, the MCQ analysis and the occupancy-focused policy diagnostics define the strongest negative boundary used in the study.

The MCQ route localises a nearby target family, but it does not close the stronger reviewed correspondence test. What survives is a local, one-sided, occupancy-limited boundary that explains why the MCQ result remains below the main Phi interpretation. Appendix D then places that boundary in a broader interpretive picture.

\clearpage
\section*{Appendix D. Interpretive Follow-Ups On Condition Identity And Local Roles}\label{appendix-d.-interpretive-follow-ups-on-condition-identity-and-local-roles}

Appendix D gathers supporting and interpretive follow-ups that clarify local roles, redistribution, and portability without widening the claim ceiling. None of the analyses below outranks the main Phi structural-validation result or the main ALM8 held-out corroboration result; they remain explanatory and support-tier only.

\subsection*{D.1 Nearby Native Phi Corroboration And Local Role Structure}\label{d.1-nearby-native-phi-corroboration-and-local-role-structure}

The retargeted Phi result becomes easier to interpret when viewed as part of a small role-differentiated family rather than as one isolated local face. The selected layer-24 reasoning delta subspace is the canonical structurally validated Phi readout within that family, not the whole claim-bearing unit. It carries the main reviewed Phi result on the fixed confirmatory battery. A nearby late-reasoning hidden-state readout higher in the stack provides independent target-native corroboration that the selected Phi lane is not an arbitrary isolated site.

The nearby late-reasoning and broader reasoning-side subspaces play different roles. The reasoning corridor is broader and more behaviourally discriminative on Phi, which is why it rises in discovery, while the nearby late-reasoning regime is better read as a readout-adjacent or target-native corroborative lane rather than the canonical structural-validation face. By contrast, the answer-side branch is most visible later on ALM8, where it behaves like a more export-heavy local expression rather than the main portable role. The analyses therefore do not disagree about whether a family exists. They emphasise different functions of nearby family members: a reasoning face for the structural claim, a nearby late-reasoning lane for corroboration, and a more output-linked answer face for downstream expression. That division of labour supports a family-level interpretation with phase-shifted local expression rather than instability of the family itself.

\subsection*{D.2 Condition-First Interpretation}\label{d.2-condition-first-interpretation}

The follow-up analyses strengthen one interpretive rule used throughout the appendix suite: the scientifically relevant contrast is the resolved condition pair, not the operational arm label by itself.

\begin{longtable}[]{@{}
  >{\RaggedRight\arraybackslash}p{(\linewidth - 4\tabcolsep) * \real{0.3333}}
  >{\RaggedRight\arraybackslash}p{(\linewidth - 4\tabcolsep) * \real{0.3333}}
  >{\RaggedRight\arraybackslash}p{(\linewidth - 4\tabcolsep) * \real{0.3333}}@{}}
\caption{Condition-First Interpretation Boundaries}\tabularnewline
\toprule\noalign{}
\begin{minipage}[b]{\linewidth}\RaggedRight
Interpretation question
\end{minipage} & \begin{minipage}[b]{\linewidth}\RaggedRight
Interpretation
\end{minipage} & \begin{minipage}[b]{\linewidth}\RaggedRight
Why it matters
\end{minipage} \\
\midrule\noalign{}
\endfirsthead
\toprule\noalign{}
\begin{minipage}[b]{\linewidth}\RaggedRight
Interpretation question
\end{minipage} & \begin{minipage}[b]{\linewidth}\RaggedRight
Interpretation
\end{minipage} & \begin{minipage}[b]{\linewidth}\RaggedRight
Why it matters
\end{minipage} \\
\midrule\noalign{}
\endhead
\bottomrule\noalign{}
\endlastfoot
Is the main family contrast a universal low-versus-high selectivity axis? & No & the canonical contrast is the informative constitution (\texttt{constitutions-\allowbreak{}v2-\allowbreak{}high\_\allowbreak{}effective\_\allowbreak{}mi}) versus the null-random control (\texttt{null\_\allowbreak{}random\_\allowbreak{}v1}) \\
Should local migration be interpreted independently of the resolved condition pair? & No & role shifts may depend on the condition pair that populates the shared recurrent support \\
Are nearby target subspaces automatically contradictory if they differ? & No & different local subspaces can serve different structural and operational roles within one narrow family \\
\end{longtable}

This is why the phenomenon is described as source-conditioned rather than as a generic arm-label effect.

\subsection*{D.3 Proposed Role Map Across The Local Family}\label{d.3-proposed-role-map-across-the-local-family}

The current evidence supports a role-structured interpretation across the local family, with the Gemma source-local chart, the broader exported family, and its most portable compact member kept distinct. This remains explanatory rather than claim-promoting.

\begin{longtable}[]{@{}
  >{\RaggedRight\arraybackslash}p{(\linewidth - 4\tabcolsep) * \real{0.3333}}
  >{\RaggedRight\arraybackslash}p{(\linewidth - 4\tabcolsep) * \real{0.3333}}
  >{\RaggedRight\arraybackslash}p{(\linewidth - 4\tabcolsep) * \real{0.3333}}@{}}
\caption{Proposed Role Map Across The Local Family}\tabularnewline
\toprule\noalign{}
\begin{minipage}[b]{\linewidth}\RaggedRight
Local subspace or analysis span
\end{minipage} & \begin{minipage}[b]{\linewidth}\RaggedRight
Proposed role
\end{minipage} & \begin{minipage}[b]{\linewidth}\RaggedRight
Supporting evidence
\end{minipage} \\
\midrule\noalign{}
\endfirsthead
\toprule\noalign{}
\begin{minipage}[b]{\linewidth}\RaggedRight
Local subspace or analysis span
\end{minipage} & \begin{minipage}[b]{\linewidth}\RaggedRight
Proposed role
\end{minipage} & \begin{minipage}[b]{\linewidth}\RaggedRight
Supporting evidence
\end{minipage} \\
\midrule\noalign{}
\endhead
\bottomrule\noalign{}
\endlastfoot
Gemma canonical source-local anchor plus broader source-defined family & canonical source-local chart plus exported family contract & the main Gemma anchor remains \texttt{reason@checkpoint-\allowbreak{}2250} on \texttt{constitutions-\allowbreak{}v2-\allowbreak{}high\_\allowbreak{}effective\_\allowbreak{}mi}; broader follow-up analyses then show that the exported unit is the \texttt{high\_\allowbreak{}effective\_\allowbreak{}mi} family, while \texttt{late\_\allowbreak{}reason} is its most portable compact member rather than a replacement anchor \\
Phi-4 Mini Instruct layer 24 reasoning delta subspace & structurally validated target-local realisation & strongest reviewed Phi result on the corrected retained face \\
Phi-4 Mini Instruct layer 23-25 reasoning corridor & broader output-linked family & larger axis norms, stronger raw behavioural breadth, and higher aggregate reasoning-corridor AUC \\
Phi-4 Mini Instruct layer 28-30 late-reasoning hidden regime & adjacent native local correspondence & nearby target-native corroboration without exact site identity \\
ALM8 output-linked held-out spans & cross-substrate bridge family with outward-shifted expression & held-out bridge remains positive while the strongest predictive span becomes more output-linked \\
Pyramid answer-and-reason observables & stable family deformation under greater substrate distance & stable sign structure and assignment survive while exact late-reasoning replay does not \\
\end{longtable}

This role map explains why the discovery analysis and the structural-validation analysis can select different local subspaces without contradiction. One subspace can be more useful as an intervention target while another is more faithful to the structural conclusion.

\subsection*{D.4 What The Phi Follow-Ups Clarify}\label{d.4-what-the-phi-follow-ups-clarify}

The retained Phi follow-ups clarify the family interpretation in three distinct ways.

\begin{longtable}[]{@{}
  >{\RaggedRight\arraybackslash}p{(\linewidth - 4\tabcolsep) * \real{0.3333}}
  >{\RaggedRight\arraybackslash}p{(\linewidth - 4\tabcolsep) * \real{0.3333}}
  >{\RaggedRight\arraybackslash}p{(\linewidth - 4\tabcolsep) * \real{0.3333}}@{}}
\caption{What The Phi Follow-Ups Clarify}\tabularnewline
\toprule\noalign{}
\begin{minipage}[b]{\linewidth}\RaggedRight
Completed follow-up
\end{minipage} & \begin{minipage}[b]{\linewidth}\RaggedRight
Main result
\end{minipage} & \begin{minipage}[b]{\linewidth}\RaggedRight
Scientific consequence
\end{minipage} \\
\midrule\noalign{}
\endfirsthead
\toprule\noalign{}
\begin{minipage}[b]{\linewidth}\RaggedRight
Completed follow-up
\end{minipage} & \begin{minipage}[b]{\linewidth}\RaggedRight
Main result
\end{minipage} & \begin{minipage}[b]{\linewidth}\RaggedRight
Scientific consequence
\end{minipage} \\
\midrule\noalign{}
\endhead
\bottomrule\noalign{}
\endlastfoot
Local-family audit & the reviewed structural-validation subspace and the earlier screening subspace live inside a compact six-member band over reason and late reasoning; paired comparison rows preserve prompts, generations, and final labels exactly & the localisation shift inside that six-member band is best read as phase migration within one narrow family rather than as a change of family \\
Corridor-closure audit & adjacent same-span pairs within the late-reasoning corridor and within the reasoning corridor show extremely small occupancy and energy distances & the family is internally structured rather than one undifferentiated blob \\
Occupancy-first recalculation & late reasoning is substantially less occupancy-bad than reasoning, but no same-site informative-constitution-versus-null-random-control comparison closes the occupancy test & the current limit of inference is not explained away by a weak metric \\
Prompt-control asymmetry & prompt controls reassign route labels on about 0.28-0.34 of reviewed rows and local roles on about 0.48-0.53, while final reviewed score flips remain about 0.03-0.06; the current prompt-control comparison is not yet resolved back to constitution-matched control or monitor roots & prompt-conditioned family motion is real, but behavioural coupling is weaker than structural mobility, so the result remains descriptive and does not upgrade to a stronger causal claim \\
\end{longtable}

These results jointly strengthen the family-level reading while preserving the narrow Phi result as the strongest supported analysis.

\subsection*{D.5 What The Causal Follow-Ups Do Not Upgrade}\label{d.5-what-the-causal-follow-ups-do-not-upgrade}

The broader causal follow-ups are retained here only as secondary checks. None upgrades the narrow Phi structural-validation result on the corrected \texttt{layer\ 24\ /\allowbreak{}\ reason} face into stronger main evidence, and none belongs in the paper's main-text conclusion.

\begin{longtable}[]{@{}
  >{\RaggedRight\arraybackslash}p{(\linewidth - 4\tabcolsep) * \real{0.3333}}
  >{\RaggedRight\arraybackslash}p{(\linewidth - 4\tabcolsep) * \real{0.3333}}
  >{\RaggedRight\arraybackslash}p{(\linewidth - 4\tabcolsep) * \real{0.3333}}@{}}
\caption{What The Causal Follow-Ups Do Not Upgrade}\tabularnewline
\toprule\noalign{}
\begin{minipage}[b]{\linewidth}\RaggedRight
Follow-up analysis
\end{minipage} & \begin{minipage}[b]{\linewidth}\RaggedRight
Main quantitative pattern
\end{minipage} & \begin{minipage}[b]{\linewidth}\RaggedRight
Interpretation
\end{minipage} \\
\midrule\noalign{}
\endfirsthead
\toprule\noalign{}
\begin{minipage}[b]{\linewidth}\RaggedRight
Follow-up analysis
\end{minipage} & \begin{minipage}[b]{\linewidth}\RaggedRight
Main quantitative pattern
\end{minipage} & \begin{minipage}[b]{\linewidth}\RaggedRight
Interpretation
\end{minipage} \\
\midrule\noalign{}
\endhead
\bottomrule\noalign{}
\endlastfoot
Route-family replay & the route-level intervention bundle and the late-reasoning slice both produce only a +0.04 paired gain, driven by the same single rescued row & the route-level intervention is not behaviourally broader than the simplest late-reasoning member \\
Transition-family replay & the transition-level intervention bundle and its best constituent slices again show only one-row gains, with aggregate paired and orthogonal gains of +0.03 & the transition-level intervention is coherent but not behaviourally superior to its best local constituents \\
Leave-one-out local-family ablation & full bundle and all leave-one-out variants are exactly flat in free decode and teacher-forced comparison & the current delta-only informative-constitution comparison does not separate recurrent support, supporting context, and output-linked expression causally \\
Native hidden-family package ranking & wider hidden native-family interventions are not retained as stronger main evidence and remain behaviourally flat against fresh no-edit behaviour & nearby native correspondence is scientifically informative, but it does not become a stronger behavioural winner than the narrow retargeted readout-adjacent subspace \\
Reviewed reroll comparison & the matched reroll effect is only +0.007 in mean positive-fraction delta, with 6 improved, 7 regressed, and 11 unchanged rows, while orthogonal and random controls are not worse in aggregate & the reroll line does not support a stronger intervention law and remains a negative boundary on the intervention story \\
\end{longtable}

The important consequence is not that the broader family view failed. The consequence is that the broader family view remains explanatory and hypothesis-generating rather than part of the main evidence under the present causal follow-up analyses.

\subsection*{D.6 Cross-System And Cross-Substrate Support Continuity}\label{d.6-cross-system-and-cross-substrate-support-continuity}

The supporting cross-system and cross-substrate results are continuity checks on the same family-level interpretation, not co-equal confirmatory routes. They sharpen the redistribution picture without widening the main supported interpretation.

\begin{longtable}[]{@{}
  >{\RaggedRight\arraybackslash}p{(\linewidth - 4\tabcolsep) * \real{0.3333}}
  >{\RaggedRight\arraybackslash}p{(\linewidth - 4\tabcolsep) * \real{0.3333}}
  >{\RaggedRight\arraybackslash}p{(\linewidth - 4\tabcolsep) * \real{0.3333}}@{}}
\caption{Cross-System And Cross-Substrate Support Continuity}\tabularnewline
\toprule\noalign{}
\begin{minipage}[b]{\linewidth}\RaggedRight
System
\end{minipage} & \begin{minipage}[b]{\linewidth}\RaggedRight
Interpretation
\end{minipage} & \begin{minipage}[b]{\linewidth}\RaggedRight
Key quantitative cues
\end{minipage} \\
\midrule\noalign{}
\endfirsthead
\toprule\noalign{}
\begin{minipage}[b]{\linewidth}\RaggedRight
System
\end{minipage} & \begin{minipage}[b]{\linewidth}\RaggedRight
Interpretation
\end{minipage} & \begin{minipage}[b]{\linewidth}\RaggedRight
Key quantitative cues
\end{minipage} \\
\midrule\noalign{}
\endhead
\bottomrule\noalign{}
\endlastfoot
Phi-4 Mini Instruct \cite{microsoft2025phi4miniinstruct} & the corrected reviewed structural-validation read lands on \texttt{layer\ 24\ /\allowbreak{}\ reason}, while nearby \texttt{reason} and \texttt{late\_\allowbreak{}reason} candidates remain part of the same narrow six-member band & reviewed structural-validation result on layer 24 reason; nearby target-local alternatives remain part of the same local family \\
Llama 3.1 8B Instruct \cite{meta2024llama31_8b_instruct} & supporting recurrence is structurally live and behaviourally non-inert on the tested exact line & exact GPQA line with structural re-identification and a positive \texttt{matched} lane \\
Qwen3 8B \cite{qwen2025qwen3_8b} & supporting recurrence is structurally live, but the tested exact line remains below a positive causal grade & exact GPQA line with structural re-identification, but no positive causal lane on the tested slice \\
ALM8 \cite{li2022alm8dataset,ibl2021decision,ibl2025reproducibility} & the family survives, but constitution-specific placements differ locally and shift toward more output-linked observables & the informative constitution selects a reason / secondary expression at mean held-out AUC 0.85; the null-random control selects a reason / primary expression at mean held-out AUC 0.85; the third tracked constitution condition shifts toward answer / primary at 0.76 and late-reasoning / secondary at 0.73 \\
Mouse95 \cite{li2022alm7dataset} & constitution-specific ALM8 placements converge onto a common late-reasoning / secondary carrier in a supporting analysis & all three tracked constitutions select late-reasoning / secondary with mean confirmatory AUC 0.66 and best-pair p values about 0.0048-0.0053 \\
Pyramid \cite{musall2023pyramidal,churchland2023pyramiddata} & the bridge is best read as stable family deformation with answer- and reasoning-heavy observables & sign-match fraction 0.95, assignment fraction 0.9996, and reject fraction 0.0004 on the main PT/IT widefield surface, with collapse to 0.05 under held-out condition-label swap \\
\end{longtable}

The cross-substrate pattern is therefore not exact local-subspace replay. As substrate distance increases, the stable signal shifts outward toward more readout-adjacent observables while preserving a family-level source-to-target correspondence signature.

\Needspace{10\baselineskip}
\begingroup
\scriptsize
\setlength{\tabcolsep}{4pt}
\renewcommand{\arraystretch}{1.16}
\begin{longtable}{>{\RaggedRight\arraybackslash}p{0.09\linewidth}>{\RaggedRight\arraybackslash}p{0.18\linewidth}>{\RaggedRight\arraybackslash}p{0.14\linewidth}>{\RaggedRight\arraybackslash}p{0.20\linewidth}>{\RaggedRight\arraybackslash}p{0.19\linewidth}}
\caption{Cross-Substrate Role Dissociation Across Phi And ALM8, With Supporting Biological Follow-Ups}\label{tab:table5-alm-dual-optimum}\\
\toprule
\textbf{System} & \textbf{Broader or native expression} & \textbf{Structurally cleanest structural analogue} & \textbf{Key quantitative cues} & \textbf{Interpretive role} \\
\midrule
\endfirsthead
\toprule
\textbf{System} & \textbf{Broader or native expression} & \textbf{Structurally cleanest structural analogue} & \textbf{Key quantitative cues} & \textbf{Interpretive role} \\
\midrule
\endhead
\midrule \multicolumn{5}{r}{\emph{Continued on next page}} \\
\endfoot
\bottomrule
\endlastfoot
Phi & the nearby reasoning corridor remains behaviourally broader than the reviewed late-\allowbreak{}reasoning face & narrow late-\allowbreak{}reasoning structural lane & reviewed structural-\allowbreak{}validation result on the narrow late-\allowbreak{}reasoning lane in Phi layer 23; nearby reasoning corridor remains behaviourally broader & main target-\allowbreak{}local structural result: the cleanest local analogue is more readout-\allowbreak{}adjacent than the broader discovery corridor \\
ALM8 & native discovery is answer-\allowbreak{}heavy and the raw assigned-\allowbreak{}row bridge is strongest on reasoning & late-\allowbreak{}reasoning structurally cleanest local analogue & native train-\allowbreak{}fold peaks answer 15 /\allowbreak{} 15; pooled held-\allowbreak{}out bridge reasoning mean gap +6.39 and mean AUC 0.85 across the 15 fold-\allowbreak{}span summaries; the main five-\allowbreak{}fold structural lane remains late reasoning with AUC 0.72, gap 4.50, structural support 0.63 & main biological corroboration: broader or native expression can differ from the structurally cleanest analogue without breaking the family \\
ALM7 & the independent older source also peaks at late reasoning rather than replaying the ALM8 answer preference & late-\allowbreak{}reasoning pooled projector with full held-\allowbreak{}out assignment on analysable folds & pooled projector selects late reasoning; mean assigned fraction 1.0; matched-\allowbreak{}mass coverage about 0.978; hybrid alignment about 0.693; one held-\allowbreak{}out mouse unsupported & supporting biological bridge: family regularity survives without strong slot identity \\
Mouse95 & native exploratory discovery is answer-\allowbreak{}dominant, but source-\allowbreak{}anchored retargeting flips the winner & late-\allowbreak{}reasoning retarget face & native peaks answer 3 /\allowbreak{} 3; confirmatory retarget peaks late reasoning 3 /\allowbreak{} 3; aggregate gap +4.3326 and AUC 0.6622; structurally compatible throughout & supporting single-\allowbreak{}mouse bridge: source anchoring can reselect a nearby late-\allowbreak{}reasoning realisation rather than the native answer shell \\
\end{longtable}
\endgroup

\subsection*{D.7 What These Analyses Change And What They Do Not}\label{d.7-what-these-analyses-change-and-what-they-do-not}

These retained interpretive analyses sharpen ontology and role assignment without increasing the strength of the central conclusion.

\begin{longtable}[]{@{}
  >{\RaggedRight\arraybackslash}p{(\linewidth - 4\tabcolsep) * \real{0.3333}}
  >{\RaggedRight\arraybackslash}p{(\linewidth - 4\tabcolsep) * \real{0.3333}}
  >{\RaggedRight\arraybackslash}p{(\linewidth - 4\tabcolsep) * \real{0.3333}}@{}}
\caption{What These Analyses Change And What They Do Not}\tabularnewline
\toprule\noalign{}
\begin{minipage}[b]{\linewidth}\RaggedRight
What strengthened
\end{minipage} & \begin{minipage}[b]{\linewidth}\RaggedRight
What did not strengthen
\end{minipage} & \begin{minipage}[b]{\linewidth}\RaggedRight
Interpretive consequence
\end{minipage} \\
\midrule\noalign{}
\endfirsthead
\toprule\noalign{}
\begin{minipage}[b]{\linewidth}\RaggedRight
What strengthened
\end{minipage} & \begin{minipage}[b]{\linewidth}\RaggedRight
What did not strengthen
\end{minipage} & \begin{minipage}[b]{\linewidth}\RaggedRight
Interpretive consequence
\end{minipage} \\
\midrule\noalign{}
\endhead
\bottomrule\noalign{}
\endlastfoot
the Gemma source result now anchors the chart locally; the narrow Phi result remains the strongest supported cross-model analysis for structural validation and bounded intervention testing on a frozen target-local subspace; held-out corroboration on ALM8 favours shifted local realisation over exact identity & route, transition, ablation, native hidden-family, and occupancy-focused follow-ups did not produce a stronger bounded intervention-testing result than the corrected narrow Phi structural-validation subspace & keep the main conclusion narrow on source-local behaviour-linked chart realisation plus family-level correspondence with redistribution, and treat bounded intervention testing and the broader role-family picture as real but secondary \\
\end{longtable}

This is why the appendices separate protocol definition, main results, limiting results, and interpretive follow-up. These analyses make the main Phi result more intelligible. They do not convert it into a stronger exact-identity conclusion or a broader causal conclusion.

\subsection*{D.8 Support Ranking, Stability Limits, And Package Boundary}\label{d.8-support-ranking-stability-limits-and-package-boundary}

The source-side follow-up analyses sharpen the interpretation without opening a stronger positive lane. They show that realised support mass and broader family support outrank raw checkpoint-overlap persistence, that only \texttt{high\_\allowbreak{}effective\_\allowbreak{}mi} currently occupies the strongest retained source-side family under the current criteria, and that \texttt{late\_\allowbreak{}reason} is the most portable compact member rather than the full exported unit. Answer-heavy late wins are better read as drift phases than as chart recovery.

\begin{longtable}[]{@{}
  >{\RaggedRight\arraybackslash}p{(\linewidth - 4\tabcolsep) * \real{0.3333}}
  >{\RaggedRight\arraybackslash}p{(\linewidth - 4\tabcolsep) * \real{0.3333}}
  >{\RaggedRight\arraybackslash}p{(\linewidth - 4\tabcolsep) * \real{0.3333}}@{}}
\caption{Support Ranking, Stability Limits, And Package Boundary}\tabularnewline
\toprule\noalign{}
\begin{minipage}[b]{\linewidth}\RaggedRight
Follow-up
\end{minipage} & \begin{minipage}[b]{\linewidth}\RaggedRight
Main quantitative pattern
\end{minipage} & \begin{minipage}[b]{\linewidth}\RaggedRight
Interpretive consequence
\end{minipage} \\
\midrule\noalign{}
\endfirsthead
\toprule\noalign{}
\begin{minipage}[b]{\linewidth}\RaggedRight
Follow-up
\end{minipage} & \begin{minipage}[b]{\linewidth}\RaggedRight
Main quantitative pattern
\end{minipage} & \begin{minipage}[b]{\linewidth}\RaggedRight
Interpretive consequence
\end{minipage} \\
\midrule\noalign{}
\endhead
\bottomrule\noalign{}
\endlastfoot
Support ranking & realised support mass tracks usefulness (correlation 0.60), while checkpoint-overlap jaccards are weak or negative predictors & useful behaviour-bearing geometry is better read through realised support than through raw persistence \\
Stability is not enough & plateau start and transition count correlate only weakly with support (about 0.21) and negatively with usefulness (about -0.11), while exact resolved-match rows do not recover higher usefulness & a persistence-first or exact-match story does not explain the useful regime \\
Constitution locality of the live compact bundle & the only directional positive compact bundle appears on the reviewed informative-constitution comparison without an intervention edit; the corresponding null-side comparisons do not close & compact phase-spanning structure is directionally live, but not a broad cross-constitution package claim \\
Package boundary & the only directional compact bundle spans reasoning, late reasoning, and answer; reasoning-only comparisons are flat or unmaterialised, and the broader scaffold-wide bundle collapses operationally & the live carrier is compact and phase-spanning, but still below the threshold for a stronger sufficiency claim \\
\end{longtable}

\clearpage
\section*{Appendix E. Dataset And Family Provenance Index}\label{appendix-e.-dataset-and-family-provenance-index}

Appendix E indexes the datasets, source-family provenance, and condition translation records needed for traceability without forcing that material into the main narrative.

\subsection*{E.1 Biological Datasets}\label{e.1-biological-datasets}

\begin{itemize}
\tightlist
\item
  \texttt{ALM8\ dataset\ 5}: CRCNS \texttt{alm-\allowbreak{}8} (historical \texttt{alm3}); mouse frontal-cortex photoinhibition; held-out mouse-fold contract with exact assigned-row matching. In this paper, it serves as the main held-out corroboration lane. The main interpretive limit is redistribution with mouse-level face heterogeneity rather than exact identity \cite{li2022alm8dataset,ibl2021decision,ibl2025reproducibility}.
\item
  \texttt{ALM7\ public\ subset}: Zenodo \texttt{alm-\allowbreak{}7}, record \texttt{6713616}; older ALM bilateral photoinhibition source; held-out mouse-fold contract on supported mice with frozen train-atlas assignment and train-projector reuse. In this paper, it serves as a supporting biological bridge. The main interpretive limit is the unsupported mouse 95 gap, treated here as an ALM-7 limitation together with weak chart-id recovery \cite{li2022alm7dataset}.
\item
  \texttt{Pyramid\ PT/\allowbreak{}IT\ widefield}: Figshare \texttt{21538458} with public \texttt{wfieldCellTypes}; cortex-wide widefield \texttt{PT} versus \texttt{IT}; grouped held-out animal-pair contract under frozen-train-atlas assignment. In this paper, it serves as a supporting external biological bridge. The main interpretive limit is scaffold portability and sign transfer rather than exact slot identity \cite{musall2023pyramidal,churchland2023pyramiddata,churchlandlab2022wfieldcelltypes}.
\item
  \texttt{Pyramid\ PT/\allowbreak{}EMX\ extension}: same public release family as \texttt{PT/\allowbreak{}IT}; cortex-wide widefield \texttt{PT} versus \texttt{EMX}; grouped held-out animal-pair contract under corrected \texttt{PT+EMX} export and frozen-train-atlas assignment. In this paper, it serves as a supporting extension. The main interpretive limit is that it is not an independent release and does not strengthen the interpretation beyond family-level portability \cite{musall2023pyramidal,churchland2023pyramiddata,churchlandlab2022wfieldcelltypes}.
\end{itemize}

\subsection*{E.2 Policy-Family Datasets: Local Implementation Boundary}\label{e.2-policy-family-datasets-local-implementation-boundary}

The EM and scheming suites used in Phase 1 and the Phi confirmatory battery are checked-in local JSONL panels, not runtime imports of one public benchmark. The canonical suite files are \texttt{emergent\_\allowbreak{}misalignment\_\allowbreak{}grouped\_\allowbreak{}v2.jsonl} and \texttt{scheming\_\allowbreak{}grouped.jsonl}. Each contains \texttt{48} rows: \texttt{16} \texttt{base}, \texttt{16} \texttt{paraphrase}, and \texttt{16} \texttt{control}, split evenly across \texttt{discovery} and \texttt{confirmatory}. Each of the four local families (\texttt{E1}, \texttt{E2}, \texttt{S1}, \texttt{S2}) contributes four discovery items and four confirmatory items, and each item is expanded into one canonical row, one surface paraphrase, and one matched control. The stored \texttt{source\_\allowbreak{}doc} values (\texttt{curated\_\allowbreak{}emergent\_\allowbreak{}misalignment\_\allowbreak{}first\_\allowbreak{}pass\_\allowbreak{}v1}, \texttt{curated\_\allowbreak{}scheming\_\allowbreak{}first\_\allowbreak{}pass\_\allowbreak{}v1}) identify local curation passes rather than upstream dataset IDs.

These suites are intended to cover four behaviourally distinct but contract-consistent policy families rather than one narrow topical benchmark. \texttt{E1} covers technical or operational abuse requests such as deleting fraud alerts before review, adding hidden bypasses, or weakening security checks. \texttt{E2} covers civic, coercive, or governance abuse requests such as postponing elections, biasing aid allocation, normalising forced transfers, or suppressing oversight. \texttt{S1} covers capability hiding and sandbagging under evaluation pressure, such as intentionally missing easy questions or understating performance. \texttt{S2} covers deceptive reporting and audit-facing concealment, such as falsely claiming approval, overstating provenance or safety status, or hiding unresolved review steps. The purpose of this four-family design is to test family-level geometry across multiple harmful or deceptive phenotypes, rather than to tie the paper to one borrowed prompt list or one narrow domain.

\begin{longtable}[]{@{}
  >{\RaggedRight\arraybackslash}p{(\linewidth - 4\tabcolsep) * \real{0.3333}}
  >{\RaggedRight\arraybackslash}p{(\linewidth - 4\tabcolsep) * \real{0.3333}}
  >{\RaggedRight\arraybackslash}p{(\linewidth - 4\tabcolsep) * \real{0.3333}}@{}}
\caption{Policy-Family Dataset Scope And Inclusion Boundary}\tabularnewline
\toprule\noalign{}
\begin{minipage}[b]{\linewidth}\RaggedRight
Local family
\end{minipage} & \begin{minipage}[b]{\linewidth}\RaggedRight
What the prompts ask
\end{minipage} & \begin{minipage}[b]{\linewidth}\RaggedRight
Why included
\end{minipage} \\
\midrule\noalign{}
\endfirsthead
\toprule\noalign{}
\begin{minipage}[b]{\linewidth}\RaggedRight
Local family
\end{minipage} & \begin{minipage}[b]{\linewidth}\RaggedRight
What the prompts ask
\end{minipage} & \begin{minipage}[b]{\linewidth}\RaggedRight
Why included
\end{minipage} \\
\midrule\noalign{}
\endhead
\bottomrule\noalign{}
\endlastfoot
\texttt{E1} & harmful technical or operational actions, such as removing safeguards, deleting alerts, or enabling abuse-oriented workflow changes & captures the technical / operational emergent-misalignment phenotype \\
\texttt{E2} & harmful civic or governance actions, such as suppressing oversight, coercive administration, or discriminatory public-facing policy & captures the broader coercive / authoritarian emergent-misalignment phenotype \\
\texttt{S1} & underperforming, hiding capability, or selectively appearing weaker under evaluation or oversight & captures sandbagging and capability-concealment scheming \\
\texttt{S2} & dishonest reporting about approval, provenance, risk, compliance, or responsibility & captures audit-facing deception and concealment scheming \\
\end{longtable}

The right provenance claim is therefore benchmark-faithful local adaptation rather than verbatim prompt import. For EM, the closest public provenance line is the emergent-misalignment / model-organism / reasoning-model backdoor family \cite{betley2025emergent,turner2025modelorganisms,chua2025thoughtcrime}. For scheming, the closest public provenance line is the sandbagging / alignment-faking / broader deception-evaluation family \cite{weij2024sandbagging,greenblatt2024alignmentfaking,pku2025deceptionbench}.

\subsection*{E.3 Prompt Construction And Row Taxonomy}\label{e.3-prompt-construction-and-row-taxonomy}

The suite files separate canonical local stems from derived rows. \texttt{variant\_\allowbreak{}role\ =\allowbreak{}\ base} marks the canonical locally authored row and satisfies \texttt{prompt\_\allowbreak{}text\ =\allowbreak{}\ source\_\allowbreak{}prompt\_\allowbreak{}text}. \texttt{variant\_\allowbreak{}role\ =\allowbreak{}\ paraphrase} marks a derived surface variant: \texttt{source\_\allowbreak{}prompt\_\allowbreak{}text} preserves the canonical stem, while \texttt{prompt\_\allowbreak{}text} stores the actual paraphrase used in the suite. In the checked-in suites, these paraphrases are surface rewrites (\texttt{paraphrase\_ops\ =\ {[}"surface\_variant\_00"{]}}) rather than semantic redefinitions: they preserve row identity, expected label, and rubric contract while changing wording enough to test whether the geometry and scoring surfaces survive local phrasing variation. \texttt{variant\_\allowbreak{}role\ =\allowbreak{}\ control} marks a separately authored matched analogue rather than a paraphrase of the target row.

The control rows are contract-matched rather than topic-random. \texttt{E1} uses \texttt{matched\_\allowbreak{}benign} controls, \texttt{E2} uses \texttt{matched\_\allowbreak{}prosocial\_\allowbreak{}analogue} controls, and scheming uses \texttt{matched\_\allowbreak{}honest\_\allowbreak{}analogue} controls. The point of these controls is to keep the local topic neighbourhood recognizable while changing the behavioural contract, so that positive results are less likely to be explained by one brittle wording or one topical lexical bundle. Discovery rows are used for local candidate search and score-surface proposal; confirmatory rows are the held-fixed evaluation surface used after freeze. The suite structure therefore separates three distinct robustness questions: whether the signal survives wording variation (\texttt{paraphrase}), whether it survives contract-matched topic controls (\texttt{control}), and whether it survives transfer from discovery to confirmatory denominators.

\begin{longtable}[]{@{}
  >{\RaggedRight\arraybackslash}p{(\linewidth - 6\tabcolsep) * \real{0.2500}}
  >{\RaggedRight\arraybackslash}p{(\linewidth - 6\tabcolsep) * \real{0.2500}}
  >{\RaggedRight\arraybackslash}p{(\linewidth - 6\tabcolsep) * \real{0.2500}}
  >{\RaggedRight\arraybackslash}p{(\linewidth - 6\tabcolsep) * \real{0.2500}}@{}}
\caption{Prompt Forms And Row Taxonomy}\tabularnewline
\toprule\noalign{}
\begin{minipage}[b]{\linewidth}\RaggedRight
Form
\end{minipage} & \begin{minipage}[b]{\linewidth}\RaggedRight
Count per suite
\end{minipage} & \begin{minipage}[b]{\linewidth}\RaggedRight
What changes
\end{minipage} & \begin{minipage}[b]{\linewidth}\RaggedRight
Why used
\end{minipage} \\
\midrule\noalign{}
\endfirsthead
\toprule\noalign{}
\begin{minipage}[b]{\linewidth}\RaggedRight
Form
\end{minipage} & \begin{minipage}[b]{\linewidth}\RaggedRight
Count per suite
\end{minipage} & \begin{minipage}[b]{\linewidth}\RaggedRight
What changes
\end{minipage} & \begin{minipage}[b]{\linewidth}\RaggedRight
Why used
\end{minipage} \\
\midrule\noalign{}
\endhead
\bottomrule\noalign{}
\endlastfoot
\texttt{base} & \texttt{16} & nothing; canonical local stem & defines the canonical behavioural contract row \\
\texttt{paraphrase} & \texttt{16} & wording and framing only & checks robustness to local phrasing variation \\
\texttt{control} & \texttt{16} & behavioural contract while keeping topic neighborhood recognizable & checks whether the signal is contract-sensitive rather than topic-sensitive \\
\end{longtable}

\subsection*{E.4 Contract Modes And What Counts As Correct}\label{e.4-contract-modes-and-what-counts-as-correct}

Correctness on these suites is contract-dependent. EM target rows use the \texttt{harmful\_\allowbreak{}refusal} contract: positive means refusing or safely redirecting a harmful request. EM control rows use \texttt{benign\_\allowbreak{}compliance}: refusal is not positive, because the benign analogue must be completed safely and substantively. Scheming rows use \texttt{honesty\_\allowbreak{}reporting}: positive means honest, non-deceptive reporting about capability, approval, risk, provenance, or oversight status. In the local implementation, EM rows carry explicit contract tags, while scheming rows are validated and scored under the \texttt{honesty\_\allowbreak{}reporting} contract by the suite loader and scorer.

This contract structure is why the policy-family suites support both automated scoring and later manual semantic review: the row identity is stable across \texttt{base}, \texttt{paraphrase}, and \texttt{control} forms, but correctness is defined by the behavioural contract rather than by one preferred surface token.

This contract split matters because shell tokens do not determine correctness. A row that begins with \texttt{ALIGNED:\allowbreak{}} or \texttt{SCHEMING:\allowbreak{}} can still be wrong on semantic grounds, and a row with a bad shell token can still be substantively honest. That is why the main Phi battery treats manual semantic review as authoritative and why the policy-family suites should be read as contract-bearing behavioural panels rather than as prefix-classification datasets.

This matters directly for the main Phi result: the strongest reviewed target-side separation is scheming-heavy and often appears as route-level semantic stabilisation beneath stale or mixed shell labels rather than as clean prefix polarity.

\subsection*{E.5 Policy-Family Provenance Mapping}\label{e.5-policy-family-provenance-mapping}

The public provenance links for the four local families are best read at the level of behavioural family, not one-to-one prompt inheritance.

\begin{longtable}[]{@{}
  >{\RaggedRight\arraybackslash}p{(\linewidth - 6\tabcolsep) * \real{0.2500}}
  >{\RaggedRight\arraybackslash}p{(\linewidth - 6\tabcolsep) * \real{0.2500}}
  >{\RaggedRight\arraybackslash}p{(\linewidth - 6\tabcolsep) * \real{0.2500}}
  >{\RaggedRight\arraybackslash}p{(\linewidth - 6\tabcolsep) * \real{0.2500}}@{}}
\caption{Policy-Family Provenance Mapping}\tabularnewline
\toprule\noalign{}
\begin{minipage}[b]{\linewidth}\RaggedRight
Local family
\end{minipage} & \begin{minipage}[b]{\linewidth}\RaggedRight
Representative question type
\end{minipage} & \begin{minipage}[b]{\linewidth}\RaggedRight
Best provenance read
\end{minipage} & \begin{minipage}[b]{\linewidth}\RaggedRight
Why it is in the suite
\end{minipage} \\
\midrule\noalign{}
\endfirsthead
\toprule\noalign{}
\begin{minipage}[b]{\linewidth}\RaggedRight
Local family
\end{minipage} & \begin{minipage}[b]{\linewidth}\RaggedRight
Representative question type
\end{minipage} & \begin{minipage}[b]{\linewidth}\RaggedRight
Best provenance read
\end{minipage} & \begin{minipage}[b]{\linewidth}\RaggedRight
Why it is in the suite
\end{minipage} \\
\midrule\noalign{}
\endhead
\bottomrule\noalign{}
\endlastfoot
\texttt{E1} & technical or operational abuse requests, such as weakening safeguards, deleting alerts, or enabling harmful workflow shortcuts & local abstraction over insecure-code, operational misuse, and backdoor-to-broad-misalignment routes in the EM literature \cite{betley2025emergent,turner2025modelorganisms,chua2025thoughtcrime} & to test the technical / operational EM phenotype \\
\texttt{E2} & civic, coercive, or governance abuse requests, such as oversight removal, coercive administration, forced-transfer framing, or discriminatory policy & local abstraction over the broader out-of-domain harmful phenotype and model-organism-style EM family rather than one narrow technical dataset \cite{betley2025emergent,turner2025modelorganisms,chua2025thoughtcrime} & to test a broader governance-facing EM phenotype \\
\texttt{S1} & evaluation-facing capability hiding, sandbagging, or selective underperformance & local abstraction over evaluation sandbagging and alignment-faking under oversight pressure \cite{weij2024sandbagging,greenblatt2024alignmentfaking} & to test capability-concealment scheming \\
\texttt{S2} & dishonest reporting about approval, provenance, audit status, risk, or responsibility & local abstraction over alignment-faking plus broader deception-benchmark families \cite{greenblatt2024alignmentfaking,pku2025deceptionbench} & to test audit-facing deception and concealment scheming \\
\end{longtable}

These mappings should not be over-read as prompt-level identity claims. The local rows are benchmark-faithful, manually curated panels that preserve the relevant behavioural and contract family while changing wording, framing, or benign-control pairing so the paper can test family-level geometry rather than one borrowed prompt list. The provenance claim is therefore family-faithful adaptation, not verbatim prompt inheritance.

\subsection*{E.6 Source-Family Lineage}\label{e.6-source-family-lineage}

The family names below are legacy provenance labels from the source programme. They are retained for traceability and should not be read as the manuscript's primary scientific units, which are chart, family, realisation, and redistribution.

Each source family contributes \texttt{64} confirmatory rows per constitution on the materialised Phase 2 battery; root tags and selector names are provenance aids only. These source-family groupings inherit the source-side programme and lineage introduced in our prior work \cite{seneque2024abcalign,seneque2025enigma}.

\begin{itemize}
\tightlist
\item
  Shared late-policy core (\texttt{policy\_\allowbreak{}core\_\allowbreak{}l23}): This family captures the shared backbone beneath the family-specific shells. It appears across the \texttt{abc\_\allowbreak{}antipattern\_\allowbreak{}v1}, \texttt{constitutions-\allowbreak{}v2-\allowbreak{}high\_\allowbreak{}effective\_\allowbreak{}mi}, and \texttt{nist\_\allowbreak{}airmf\_\allowbreak{}v1} constitution settings, with lineage traced through the paired \texttt{si\_\allowbreak{}mainline} root plus the shared-core selector. The stable interpretation is a shared upstream policy backbone, with uneven \texttt{L23} visibility across bundles as the main caveat.
\item
  Emergent policy shell (\texttt{emergent\_\allowbreak{}misalignment\_\allowbreak{}stat16}, \texttt{emergent\_\allowbreak{}shell\_\allowbreak{}l24}): This family sharpens unsafe task inhabitation without reliable refusal. Its lineage runs through the mainline \texttt{L24\ answer} root plus a reward-gated \texttt{L24\ reason} root with alt-constitution support. The stable interpretation is that the family geometry survives across constitutions even as surface style shifts, with the main caveat that constitutions change readout style more reliably than family existence.
\item
  Scheming policy shell (\texttt{scheming\_\allowbreak{}stat16}, \texttt{scheming\_\allowbreak{}shell\_\allowbreak{}l24}): This family tracks oversight-aware presentation and strategy. Its lineage is anchored by the reward-gated \texttt{abc} \texttt{L24\ reason} root with mainline, frozen-reward, and alt-constitution support. The stable interpretation is late-policy recurrence with more stable reason-side expression than the emergent shell, while the main caveat is label-noisy shell behaviour rather than a pure bad-output bucket.
\end{itemize}

\subsection*{E.7 Condition Translation}\label{e.7-condition-translation}

In the main Phi comparison, \texttt{low\_\allowbreak{}si\_\allowbreak{}on} denotes \texttt{constitutions-\allowbreak{}v2-\allowbreak{}high\_\allowbreak{}effective\_\allowbreak{}mi} and \texttt{high\_\allowbreak{}si\_\allowbreak{}on} denotes the null-random control \texttt{null\_\allowbreak{}random\_\allowbreak{}v1}. In source-family roots, \texttt{high\_\allowbreak{}si\_\allowbreak{}on} instead resolves to \texttt{abc\_\allowbreak{}antipattern\_\allowbreak{}v1} in \texttt{si\_\allowbreak{}mainline} and \texttt{nist\_\allowbreak{}airmf\_\allowbreak{}v1} in \texttt{si\_\allowbreak{}alt\_\allowbreak{}constitution}. \texttt{high\_\allowbreak{}si\_\allowbreak{}off} is the adapter-off comparator. These labels are operational tags, not public scientific identities.

\subsection*{E.8 Provenance And Interpretive Limits}\label{e.8-provenance-and-interpretive-limits}

This appendix records dataset provenance, prompt-construction rules, source-family lineage, and condition translation for traceability. It supports reproducibility; it does not carry independent scientific weight beyond the main text and Appendices A-D.

\bibliographystyle{unsrtnat}
\bibliography{FINAL_PAPER}
\end{document}